\pdfoutput=1
\documentclass{opt2026} %

\title[ZEMA: confidence-calibrated signed descent with only one state.]{ZEMA: confidence-calibrated signed descent with only one state.}

\usepackage{booktabs}
\usepackage{multirow}

\newcommand{\EMA}{\text{\fontfamily{cmtt}\fontseries{m}\selectfont EMA}}
\newcommand{\argmin}[1]{\operatorname*{arg\,min}_{#1}}
\newcommand{\ZEMA}{\textsc{ZEMA}}

\optauthor{%
 \Name{Ali Dadsetan} \Email{Ali.Dadsetan@dal.ca}\\
 \Name{Domenic Rosati} \Email{Domenic.Rosati@dal.ca}\\
 \Name{Frank Rudzicz} \Email{Frank@dal.ca}\\
 \addr Dalhousie University and Vector Institute%
}

\begin{document}

\maketitle

\begin{abstract}%

Recent work characterizes the Adam optimizer as sign descent modulated by gradient uncertainty: the update is the sign of the batch gradient, scaled by how strongly the batch supports that sign. We apply this view directly and introduce \ZEMA{} (Z-score Evidence Momentum Adaptation), which measures that support with a per-coordinate $z$-score computed from the per-example gradients in the current batch, converts it into a sign probability, and smooths the resulting update with a single exponential moving average. Because the probability is formed before smoothing rather than after, \ZEMA{} keeps one state per parameter instead of Adam's two, at a $4.6\%$ increase in per-step time; the per-example variance it requires can also be estimated from a small fraction of the batch. In fine-tuning and pretraining experiments, \ZEMA{} is competitive with Adam, reduces the variability of the validation loss, and in some settings recovers the sign of the full-dataset gradient more consistently. We close with two directions this opens: extending confidence adaptation beyond the sign geometry, to methods such as Muon, and to differentially private training.

\end{abstract}

\section{Introduction}

Sign descent dynamics explain a considerable part of Adam's success \cite{bernstein_signsgd_2018,kunstner_noise_2023,kunstner_scaling_2025,tomihari_understanding_2025,zhao_deconstructing_2025}, alongside recent geometric accounts that are outside the scope of this paper \citep{xie_adam_2025, bernstein_old_2024}. However, sign descent methods degrade at smaller batch sizes relative to Adam \cite{semenov_benchmarking_2025,kunstner_noise_2023}, which suggests a greater sensitivity to gradient noise: even extensively tuned Signum underperforms Adam \cite{orvieto_search_2025}. 

\paragraph{Adam as uncertainty-aware adaptation} One explanation for this gap is that Adam scales its step by the uncertainty in the sign. If the sign of the stochastic batch gradient, $m_i = B^{-1} \sum_n g_i^{n}$ for per-example gradients $g_i^n$ with respect to the $i$th parameter, is positive with probability $\rho_i$, then the $i$th component of the Adam update is approximately $2\rho_i - 1$\footnote{For readers unfamiliar with this theory, see our review in Appendix~\ref{app:background-adam-as-uncertainty-aware-sign}.}. When the sign is deterministic, i.e., $\rho_i = 0$ or $\rho_i = 1$, this reduces to sign descent; when both signs are equally likely, i.e., $\rho_i = 0.5$, the update is zero. Prior work interprets Adam as a three-stage procedure for estimating $\rho_i$: (i) assume a normal distribution for the stochastic batch gradient \cite{balles_dissecting_2020,orvieto_search_2025}; (ii) estimate its mean and standard deviation, $m_i$ and $\sigma_i$, by variational inference, implemented as an exponential moving average (EMA) \cite{orvieto_search_2025}; and (iii) compute the update $2\rho_i-1$ from these estimates using an approximation to the error function, since $2\rho_i-1=\operatorname{erf}({z_i}/{\sqrt{2}})=\operatorname{erf}(m_i/(\sqrt{2}\sigma_i))$ for the $z$-score $z_i = m_i/\sigma_i$ \cite{balles_dissecting_2020}.

\paragraph{Contributions.} We take the magnitude of a signed update to reflect how strongly the batch supports that sign, and we measure that support per coordinate with a $z$-score. \ZEMA{} computes $z_i$ from the spread of the per-example gradients in the \emph{current batch}, converts it to the probability $\rho_i$ that the full-dataset gradient is positive, and smooths the resulting update with a single exponential moving average (Algorithm~\ref{alg:zema}). Adam forms this probability after smoothing, whereas \ZEMA{} forms it before, and therefore needs only one state per parameter. The per-example variance this requires costs $4.6\%$ more per step (Table~\ref{tab:fast-throughput}, Appendix~\ref{app:efficient-implementation}), and can be estimated from a small fraction of the batch (Appendix~\ref{sec:dispersion-details}). \ZEMA{} matches AdamW on LoRA fine-tuning to within $0.04$ perplexity (Table~\ref{tab:finetuning}) and improves on it by $0.6$ perplexity at 160M-parameter pretraining, at roughly a quarter of AdamW's seed-to-seed standard deviation and at the learning rate chosen for AdamW (Table~\ref{tab:pretraining}).

\paragraph{Related work} \label{sec:related}
Previous adaptive methods store only the first moment (Signum \cite{bernstein_signsgd_2018}, Lion \cite{chen2023symbolic}) or only the second (AdaGrad \cite{duchi2011adaptive}, RMSProp \cite{hinton2012rmsprop}), whereas \ZEMA{} keeps a single statistic, the $z$-score, that combines the information both provide. Muon \cite{jordan2024muon}, which works in a different geometry, and GaLore \cite{zhao2024galore}, which projects gradients into a low-rank subspace, are also used for training large language models, but their connection to signed descent is currently unclear and outside the scope of this paper; we include both in Appendix~\ref{sec:other-families} for a broad comparison. Variance-reduction strategies such as MARS \citep{yuan2024mars} could be applied on top of \ZEMA{}, which we leave to future work. Finally, ours is not the first work to use per-example variance, or to motivate Adam in terms of a $z$-score-like statistic \citep{kunze2021adaptive,balles_dissecting_2020,patitucci2026adaptive}, but these works do not use it to reduce the number of optimizer states, nor do they evaluate extensively in the LLM regime, where large-batch pretraining involves non-stationary gradients from rare tokens.

\section{ZEMA: Z-score Evidence Momentum Adaptation}

\paragraph{Adaptivity as evidence and confidence.} Adam estimates $\rho_i$ indirectly, through two moments smoothed across steps. We estimate it directly from the current batch. The batch-level \emph{evidence} is the $z$-score $z_i = m_i/\sigma_i$, which measures how far the batch mean sits from zero relative to its sampling error. Under the Gaussian assumption, the probability that the full-dataset gradient is positive is then $\rho_i = \Phi(z_i)$, where $\Phi$ is the CDF of $\mathcal{N}(0,1)$; we call this probability the \emph{confidence} in the sign. The confidence-adapted sign descent direction is $u_i = 2\rho_i - 1 = \operatorname{erf}(z_i/\sqrt{2})$, which we smooth over time to obtain the direction $\hat{u}_i$ we actually apply. Two changes relative to Adam follow: the evidence comes from statistics of the current batch rather than from moments accumulated across steps (Section~\ref{sec:zema-evidence}), and the smoothing acts on the confidence rather than on the moments it is computed from (Section~\ref{sec:zema-smoothing}). The second is what leaves \ZEMA{} with one state per parameter instead of two.

Because $\mathbb{E}[u_i]$ vanishes when the evidence does, coordinates with weak evidence take proportionally smaller steps than under Signum (Proposition~\ref{prop:damping}); and because $z_i$ grows as $\sqrt{B}$ for any fixed non-zero gradient, the step saturates to the sign of the full-dataset gradient as $B$ grows (Proposition~\ref{prop:saturation}). Adam's denominator retains the variation of the gradient across steps and so does not saturate the same way (Appendix~\ref{app:background-adam-as-uncertainty-aware-sign}); Appendix~\ref{app:zema-theory} develops the evidence view further.

\subsection{Estimating the $z$-score from batch statistics}
\label{sec:zema-evidence}
At iteration $t$ we estimate the standard error of the batch mean, $\boldsymbol{\sigma}_t$, from the per-example gradients (Algorithm~\ref{alg:zema}, line~\ref{line:zema-standard-error}). This adds computation on top of the batch gradient $\boldsymbol{m}_t$ alone, but most of it can be shared with the usual backpropagation \citep[BackPack]{dangel2019backpack}. The overhead is largest in the pretraining setting of Section~\ref{sec:pretraining}, where a direct implementation costs $1.59\times$ AdamW per step; the implementation of Appendix~\ref{app:efficient-implementation} reduces this to $1.046\times$, a $4.6\%$ increase in per-step time (Table~\ref{tab:fast-throughput}). Appendix~\ref{sec:dispersion-details} discusses a subsampling approach that reduces the cost of the dispersion further.

The evidence is then $\boldsymbol{z}_t=\boldsymbol{m}_t/(\boldsymbol{\sigma}_t+\sigma_0)$, clipped to $\pm z_{\max}$ (line~\ref{line:zema-z-score}), where $\sigma_0>0$ prevents division by a vanishing standard error. It is a property of the current batch alone, with no dependence on earlier steps.

\subsection{Smoothing the confidence over time}
\label{sec:zema-smoothing}
Two links turn $\boldsymbol{z}_t$ into a confidence. Equation~\ref{eq:rho-from-normal} gives $\boldsymbol\rho_t^{\text{CLT}} = \frac{1}{2} + \frac{1}{2} \operatorname{erf}\left({\boldsymbol{z}_t}/{\sqrt{2}}\right)$ (Algorithm~\ref{alg:zema}, line~\ref{line:zema-clt-link}), and the rational approximation of Equation~\ref{eq:erf-approximation} gives $\boldsymbol\rho_t^{\text{Adam}} = \frac{1}{2} + \frac{1}{2} \bigl({\boldsymbol{z}_t}/{\sqrt{1 + \boldsymbol{z}_t^2}}\bigr)$ (line~\ref{line:zema-adam-link}), the form in which Adam's own update can be written.

Since $\boldsymbol\rho_t$ is estimated from a single batch, we smooth it over time. Our variational inference formulation is

\begin{equation}
  \label{eq:variational-inference-for-rho}
\hat{\rho}_{t,i} = \argmin{\hat{\rho}_i} \operatorname{KL}\big(\text{Bern}(\rho_{t,i}^{\text{METHOD}}) || \text{Bern}(\hat{\rho}_i)\big) + \lambda \operatorname{KL}\big(\text{Bern}(\hat{\rho}_{t-1,i}) || \text{Bern}(\hat{\rho}_i) \big)
\end{equation} 

whose solution is available in closed form as the exponential moving average

\begin{equation}
  \label{eq:rho-ema}
\hat{\boldsymbol\rho}_{t} = \beta \hat{\boldsymbol\rho}_{t-1} + (1 - \beta) \boldsymbol\rho_{t}^{\text{METHOD}}
\end{equation}

with $\beta = \frac{\lambda}{1 + \lambda}$; a proof is given in Appendix~\ref{sec:rho-ema-proof}. The update direction is $\hat{\boldsymbol{u}}_{t} = 2 \hat{\boldsymbol\rho}_{t} - 1$. Because the smoothing and the map $\boldsymbol\rho \mapsto 2\boldsymbol\rho - 1$ commute, Algorithm~\ref{alg:zema} equivalently forms $\boldsymbol{u}_t = 2\boldsymbol\rho_t - 1$ first (lines~\ref{line:zema-clt-link} and~\ref{line:zema-adam-link}) and smooths that instead, so $\hat{\boldsymbol{u}}$ is the only state it carries.

\begin{algorithm2e}
\caption{\ZEMA{}, in its \textnormal{\textsc{CLT}} and \textnormal{\textsc{Adam-shaped}} variants.
         }
\label{alg:zema}
\SetAlgoLined
\LinesNumbered
\DontPrintSemicolon
\KwIn{step size $\eta$; smoothing $\beta \in [0,1)$; weight decay $\lambda$;
        $\sigma_0$, $z_{\max}$; link $\in \{\textnormal{\textsc{CLT}}, \textnormal{\textsc{Adam}}\}$}
\KwOut{parameters $\boldsymbol{\theta}_T$}
$\hat{\boldsymbol{u}}_{0} \leftarrow \mathbf{0}$ \tcp*{the only state, one scalar per parameter}
\For{$t \leftarrow 1$ \KwTo $T$}{
  sample a batch of size $B$; form per-example gradients $\boldsymbol{g}^{(1)}_{t},\dots,\boldsymbol{g}^{(B)}_{t}$\nllabel{line:zema-per-example-gradients}\;
  $\boldsymbol{m}_{t} \leftarrow \frac{1}{B}\sum_{n} \boldsymbol{g}^{(n)}_{t}$ \nllabel{line:zema-batch-mean}\tcp*{batch mean}
  $\boldsymbol{\sigma}_{t} \leftarrow \bigl[\frac{1}{B(B-1)}\sum_{n}(\boldsymbol{g}^{(n)}_{t}-\boldsymbol{m}_{t})^{2}\bigr]^{1/2}$
    \nllabel{line:zema-standard-error}\tcp*{standard error of $\boldsymbol{m}_t$}
  $\boldsymbol{z}_{t} \leftarrow \operatorname{clip}\bigl(\boldsymbol{m}_{t}/(\boldsymbol{\sigma}_{t}+\sigma_{0}),\ \pm z_{\max}\bigr)$
    \nllabel{line:zema-z-score}\tcp*{z-score, this batch alone}
  \uIf{\textup{link} $=$ \textnormal{\textsc{CLT}}}{
    $\boldsymbol{u}_{t} \leftarrow \operatorname{erf}(\boldsymbol{z}_{t}/\sqrt{2})$
      \nllabel{line:zema-clt-link}\tcp*{$2\boldsymbol{\rho}^{\mathrm{CLT}}_{t}-1$}
  }
  \Else{
    $\boldsymbol{u}_{t} \leftarrow \boldsymbol{z}_{t}\,(1+\boldsymbol{z}_{t}^{2})^{-1/2}$
      \nllabel{line:zema-adam-link}\tcp*{$2\boldsymbol{\rho}^{\mathrm{Adam}}_{t}-1$}
  }
  $\hat{\boldsymbol{u}}_{t} \leftarrow \beta\,\hat{\boldsymbol{u}}_{t-1} + (1-\beta)\,\boldsymbol{u}_{t}$
    \tcp*{smooth the belief, not $\boldsymbol{z}$
    }
  $\boldsymbol{\theta}_{t} \leftarrow (1-\eta\lambda)\,\boldsymbol{\theta}_{t-1} - \eta\,\hat{\boldsymbol{u}}_{t}$
    \tcp*{decoupled decay, then step}
}
\end{algorithm2e}

\section{Experiments}

We focus exclusively on LLM settings, where Adam has consistently been shown to be of particular importance \citep{kunstner_noise_2023,tomihari_understanding_2025, zhang_why_2024}. In the main paper, we present both fine-tuning and pretraining settings, to show the advantages and the limitations of \ZEMA{}; experiments on non-stationarity (rare tokens and task mixtures) are in Appendix~\ref{app:ppl-sft}. Further experiments with different optimizers, with the choice of $k$ in batch subsampling, and on computational overhead are in Appendix~\ref{app:additional-experiments}. These include a probe of how often each optimizer recovers the sign of the full-dataset gradient, where \ZEMA{} is ahead of AdamW by $9$--$12$ points on the output head in pretraining, but not separable from it on the other tensors we probed (Appendix~\ref{app:sign-estimation-accuracy}).

\subsection{Fine-tuning}
\label{sec:finetuning}

We compare both \ZEMA{} variants against AdamW on LoRA \citep{hu_lora_2022}
fine-tuning across three model scales---Llama-3.2-1B, Llama-3.2-3B and
Llama-3.1-8B---and two generation tasks, E2E NLG \citep{novikova_e2e_2017} and
SAMSum \citep{gliwa_samsum_2019}. We add Qwen3-1.7B \citep{yang_qwen3_2025} at a
single scale as a check that the result is not specific to one model family. Each optimizer is tuned on its own learning-rate grid, wide enough that
the selected rate is interior, and the selected rate is then rerun with two
further seeds. The setup, the rate-selection procedure, and the full sweeps are
in Appendix~\ref{sec:finetuning-details}.

Table~\ref{tab:finetuning} reports the best validation perplexity over epochs,
averaged over three seeds. Both optimizers see the same seeds, so each
difference is formed per seed; the reported interval is a paired $95\%$
Student-$t$ interval.

\begin{table}[t]
\centering
\caption{
\footnotesize
LoRA fine-tuning: best validation perplexity over epochs, mean of three
seeds. The first three blocks are a Llama scale ladder; Qwen3-1.7B is a
single-scale check on a second family and its perplexities are not comparable
across families. $\Delta$ is the paired difference from AdamW at the same seed, with a
$95\%$ Student-$t$ interval; positive favours AdamW. Learning rates are selected
per cell (Appendix~\ref{sec:finetuning-details}).}
\label{tab:finetuning}
\small
\begin{tabular}{llcccc}
\toprule
& & \multicolumn{2}{c}{E2E NLG} & \multicolumn{2}{c}{SAMSum} \\
\cmidrule(lr){3-4}\cmidrule(lr){5-6}
Model & Optimizer & ppl & $\Delta$ vs.\ AdamW & ppl & $\Delta$ vs.\ AdamW \\
\midrule
\multirow{3}{*}{Llama-3.2-1B} & AdamW & 2.959 & --- & 8.134 & --- \\
 & \ZEMA{}-CLT & 2.971 & $+0.011 \pm 0.038$ & 8.173 & $+0.039 \pm 0.017$ \\
 & \ZEMA{}-Adam & 2.975 & $+0.015 \pm 0.026$ & 8.172 & $+0.038 \pm 0.017$ \\
\midrule
\multirow{3}{*}{Llama-3.2-3B} & AdamW & 2.911 & --- & 7.157 & --- \\
 & \ZEMA{}-CLT & 2.932 & $+0.021 \pm 0.012$ & 7.197 & $+0.040 \pm 0.008$ \\
 & \ZEMA{}-Adam & 2.931 & $+0.020 \pm 0.011$ & 7.194 & $+0.037 \pm 0.007$ \\
\midrule
\multirow{3}{*}{Llama-3.1-8B} & AdamW & 2.878 & --- & 6.595 & --- \\
 & \ZEMA{}-CLT & 2.889 & $+0.011 \pm 0.018$ & 6.622 & $+0.026 \pm 0.026$ \\
 & \ZEMA{}-Adam & 2.887 & $+0.009 \pm 0.017$ & 6.621 & $+0.025 \pm 0.024$ \\
\midrule
\multirow{3}{*}{Qwen3-1.7B} & AdamW & 2.953 & --- & 8.297 & --- \\
 & \ZEMA{}-CLT & 2.968 & $+0.015 \pm 0.022$ & 8.344 & $+0.047 \pm 0.006$ \\
 & \ZEMA{}-Adam & 2.967 & $+0.014 \pm 0.028$ & 8.343 & $+0.045 \pm 0.009$ \\
\bottomrule
\end{tabular}
\end{table}

\textbf{The two optimizers are separated by at most $0.05$ validation
perplexity.} Across all sixteen comparisons the gap lies between $+0.009$ and
$+0.047$ perplexity, against absolute values of $2.9$--$3.0$ on E2E and
$6.6$--$8.3$ on SAMSum, or $0.3\%$ to $0.7\%$; six of the sixteen are
indistinguishable from seed noise at the $95\%$ level. We read the fine-tuning
result as practical parity, with \ZEMA{} carrying one state variable per
parameter instead of two.

The seed-to-seed spread is also smaller for both \ZEMA{} variants in four of the
six Llama cells, most clearly on SAMSum at 8B ($0.003$ and $0.004$ perplexity
against AdamW's $0.011$); it is comparable at 3B/E2E and larger at 8B/E2E.

\subsection{Pretraining}
\label{sec:pretraining}

For pretraining, we replicate the $160$M-parameter setting of
\citet{orvieto_search_2025}: a $162.2$M-parameter decoder-only transformer
trained from scratch on $3.2$B tokens of SlimPajama at context $2048$ and
global batch $256$, three seeds per arm. AdamW and Signum run at the peak
learning rates that paper selects, $8\cdot10^{-3}$ and $10^{-3}$. The two
\ZEMA{} arms are not tuned and both run at
AdamW's.
Table~\ref{tab:pretraining} adds the three one-state coordinate-wise baselines
named in Section~\ref{sec:related}---Lion, Adagrad and RMSProp---each swept on
its own grid. Appendix~\ref{sec:pretraining-details} gives the architecture,
the schedule and the corpus, and Appendix~\ref{sec:pretraining-choices} the
rate grids and the configuration choices behind these rows.

\begin{table}[h]
\centering
\caption{
\footnotesize
Pretraining, replicating the $160$M setting of
\citet{orvieto_search_2025}: final validation perplexity after $3.2$B tokens,
mean of three seeds. The interval on $\Delta$ is a $95\%$ Student-$t$ interval on
the per-seed paired difference from AdamW; negative favours \ZEMA{}. AdamW and
Signum use the rates that paper reports; the \ZEMA{} arms copy AdamW's without a
sweep; Lion, RMSProp and Adagrad are each swept on their own grid
(Appendix~\ref{sec:pretraining-choices}). All arms use weight decay $0.1$.
Optimizers from other families are in Appendix~\ref{sec:other-families}.}
\label{tab:pretraining}
\small
\begin{tabular}{lcccc}
\toprule
Optimizer & states/param & learning rate & val.\ ppl & $\Delta$ vs.\ AdamW \\
\midrule
AdamW & $2$ & $8{\cdot}10^{-3}$ & $22.79 \pm 0.46$ & --- \\
\midrule
\ZEMA{}-CLT & $1$ & $8{\cdot}10^{-3}$ & $\mathbf{22.18 \pm 0.12}$ & $-0.61 \pm 0.55$ \\
\ZEMA{}-Adam & $1$ & $8{\cdot}10^{-3}$ & $22.36 \pm 0.22$ & $-0.43 \pm 0.36$ \\
Signum & $1$ & $1{\cdot}10^{-3}$ & $25.25 \pm 1.02$ & $+2.46 \pm 0.60$ \\
RMSProp & $1$ & $1{\cdot}10^{-3}$ & $26.71 \pm 0.11$ & $+3.93 \pm 0.39$ \\
Lion & $1$ & $1{\cdot}10^{-4}$ & $29.10 \pm 1.51$ & $+6.32 \pm 1.35$ \\
Adagrad & $1$ & $1{\cdot}10^{-2}$ & $38.31 \pm 1.23$ & $+15.53 \pm 1.10$ \\
\bottomrule
\end{tabular}
\end{table}

Both \ZEMA{} variants finish ahead of AdamW, by $0.61$ and $0.43$ validation
perplexity; both paired differences resolve at the $95\%$ level. \ZEMA{} therefore
matches or exceeds AdamW at AdamW's own learning rate, with a smaller seed
spread and one state variable per parameter instead of two. A second pretraining setting supports
reading this as parity or better: at $9.31$M parameters on TinyStories, run to
the same $3.2$B tokens, the two optimizers finish level
(Appendix~\ref{app:horizon}). Both settings show the same shape over training.
AdamW leads early in validation loss---by $21.5$ perplexity at $0.26$B tokens
here---after which \ZEMA{} catches up, takes the lead by about $1.1$B tokens,
and holds it to the end, finishing level on TinyStories and $0.61$ ahead here
(Table~\ref{tab:slim-horizon}). Which optimizer is ahead therefore depends on
where training stops, which is a caution for comparisons made at a fixed update
budget.

\ZEMA{} is the only one-state method in Table~\ref{tab:pretraining} that gives
up no validation loss for the reduction; the others trail AdamW by between
$2.5$ and $15.5$ perplexity (Appendix~\ref{sec:pretraining-choices}). Keeping
one state instead of two also halves what the optimizer stores: $84.9$M values
on the bound weights against AdamW's $169.9$M
(Appendix~\ref{sec:other-families}).

\section{Conclusion}

\paragraph{Discussion \& Limitations} \ZEMA{} trades computation for memory at little to no cost in quality: it halves the optimizer state for a few percent more time per step (Appendix~\ref{app:efficient-implementation}), and at every endpoint we measured it is at parity with AdamW or ahead of it. Which to prefer depends on which resource is scarce---\ZEMA{} when memory is, as in training a large model on limited VRAM, AdamW when step time is. Two limitations qualify this: the $z$-score is formed within a single batch, so \ZEMA{} does not accumulate the across-step gradient variation that Adam's second moment carries, though this costs no validation loss in our experiments; and \ZEMA{} converges more slowly early in pretraining, with AdamW leading until about a third of the way through our longest run (Table~\ref{tab:slim-horizon}, Appendix~\ref{app:horizon}).

\paragraph{Conclusion} We introduce \ZEMA{}, an uncertainty-aware sign-based algorithm for training trans\-former-based language models. It forms the confidence in the sign of the gradient from the statistics of the current batch rather than from moments accumulated across steps, and because that confidence is formed before smoothing, one state per parameter does the work of Adam's two. In our experiments this costs little or nothing in quality: parity with AdamW on LoRA fine-tuning, to within $0.05$ validation perplexity, and $0.61$ perplexity ahead at $160$M-parameter pretraining at AdamW's own learning rate, with a smaller seed-to-seed spread. Where the optimizer state is what limits training, dropping one of its two tensors for a few percent of step time is a favourable trade.

\paragraph{Future work} We see three directions of future work. First, our work highlights that the choices inside the confidence adaptation of sign descent itself are still open. The link function assumes a Gaussian distribution for the stochastic gradient; the Student-$t$ link noted in Appendix~\ref{app:zema-theory} is one alternative, and heavier-tailed per-example distributions \cite{pmlr-v238-battash24a} are another. The target is a choice as well: the step $2\rho_i - 1$ minimizes the mean squared error to the gradient sign (Theorem~\ref{thm:best-mse-sign-estimator}), and a different loss---a more robust one, for instance---would map the same confidence to a different step.

Second, the construction is not tied to the sign geometry. Operator-norm optimizers such as Muon \citep{jordan2024muon} and Shampoo \cite{gupta2018shampoo}, which can be read as steepest descent under a norm \citep{bernstein_old_2024}, have no built-in confidence adaptation and are in that respect closer to Signum than to Adam. Adding it requires deciding how per-coordinate evidence should scale a matrix-valued step rather than one coordinate at a time; attempts exist \cite{si2025adamuon,liu2026muon,zhang2026adam,zhang2025adagrad}, but we are not aware of a principled derivation.

Third, differential privacy. \citet{tang2024dp} show that in DP-Adam the privacy noise dominates the second moment, so that the optimizer behaves like DP-SGD unless that noise is corrected out of the denominator; from the confidence view, such a correction discards the uncertainty the noise itself introduces into the sign. A confidence formed directly from batch statistics, i.e., from the $z_i$ computed in \ZEMA{}, is a more natural target for privatization.

\bibliography{sample}

\newpage
\appendix
\section{Adam as an uncertainty-aware sign descent algorithm}
\label{app:background-adam-as-uncertainty-aware-sign}

In this section we review the approaches from previous literature that treat Adam as an uncertainty-aware sign descent algorithm, together with the proofs and derivations they rest on.

\subsection{Uncertainty adaptation of the sign descent algorithm}
\label{sec:uncertainty-adaptation}
\begin{theorem}[Best MSE sign estimator]
\label{thm:best-mse-sign-estimator}
Suppose $\mathbf{g}$ is a random variable taking values in $\mathbb{R}^d$, and suppose $\operatorname{sign}(g_i)\in\{-1,+1\}$ for each $i$. Let

$$
\bigl[\boldsymbol{\rho}\bigr]_i = \rho_i = \mathbb{P}\bigl(\operatorname{sign}(g_i)=+1\bigr).
$$

Then:
\begin{enumerate}
\item The expectation of the sign vector satisfies

\begin{equation}
    \mathbb{E}[\operatorname{sign}(\mathbf{g})]
    =
    2\boldsymbol{\rho}-1.
\end{equation}

\item Among all deterministic vectors $\boldsymbol{x}\in\mathbb{R}^d$, the vector minimizing the mean squared error to $\operatorname{sign}(g)$ is
\(
(2\rho_1-1,\ldots,2\rho_d-1).
\)
That is,
\begin{equation}
\boldsymbol{x}^\star
=
\argmin{\boldsymbol{x}\in\mathbb{R}^d}
\mathbb{E}\!\left[
\left\|\boldsymbol{x}-\operatorname{sign}(\mathbf{g})\right\|_2^2
\right] = 2\boldsymbol{\rho}-1.
\end{equation}

\end{enumerate}
\end{theorem}
 The proof is given in Appendix~\ref{sec:mse_proof} below. \footnote{We like to mention that while this theorem is essentially equivalent to ~\citet[Lemma 1]{balles_dissecting_2020}, we have reduced the clutter in the equations and removed usage of absolute value operator by defining $\rho_i$ as the probability of the sign being positive, rather than the probability of the sign being equal to the sign of the true gradient.} 

\subsection{Proof of Theorem~\ref{thm:best-mse-sign-estimator}}
While this theorem generalizes ~\citet[Lemma 1]{balles_dissecting_2020} as it does not assume that $\mathbf{g}$ is normally distributed, it is itself a special case of the well-known fact that the mean of a random variable is the unique minimizer of the mean squared error to that random variable.
\label{sec:mse_proof}
For the first statement, since $\operatorname{sign}(g_i)$ takes the value $+1$ with probability $\rho_i$ and the value $-1$ with probability $1-\rho_i$,

$$
\mathbb{E}[\operatorname{sign}(g_i)]
=
\rho_i(+1)+(1-\rho_i)(-1)
=
2\rho_i-1.
$$

Applying this argument coordinate-wise gives

$$
\mathbb{E}[\operatorname{sign}(\mathbf{g})]
=
(2\rho_1-1,\ldots,2\rho_d-1).
$$

For the second statement, let

$$
\mathbf{S}=\operatorname{sign}(\mathbf{g})
$$

and consider the objective

$$
L(\boldsymbol{x})
=
\mathbb{E}\!\left[\|\boldsymbol{x}-\mathbf{S}\|_2^2\right].
$$

Expanding the squared norm,

$$
L(\boldsymbol{x})
=
\mathbb{E}\!\left[\|\boldsymbol{x}\|_2^2
-2\boldsymbol{x}^\top \mathbf{S}
+\|\mathbf{S}\|_2^2\right].
$$

Since $\boldsymbol{x}$ is deterministic,

$$
L(\boldsymbol{x})
=
\|\boldsymbol{x}\|_2^2
-2\boldsymbol{x}^\top\mathbb{E}[\mathbf{S}]
+\mathbb{E}[\|\mathbf{S}\|_2^2].
$$

The last term does not depend on $\boldsymbol{x}$. Completing the square gives

$$
L(\boldsymbol{x})
=
\left\|\boldsymbol{x}-\mathbb{E}[\mathbf{S}]\right\|_2^2
+
\mathbb{E}[\|\mathbf{S}\|_2^2]
-
\|\mathbb{E}[\mathbf{S}]\|_2^2.
$$

Therefore the unique minimizer is

$$
\boldsymbol{x}^\star=\mathbb{E}[\mathbf{S}].
$$

Using the first part,

$$
\boldsymbol{x}^\star
=
(2\rho_1-1,\ldots,2\rho_d-1).
$$

\jmlrQED

\subsection{Distribution of sign of stochastic batch gradient}
\label{sec:distribution-of-sign-of-normal}
Assuming gradients are normally distributed, $g_i \sim \mathcal{N}(m_i, \sigma_i^2)$, the probability of the sign of $g_i$ being positive is given by
\begin{equation}
  \label{eq:rho-from-normal}
\rho_i = \mathbb{P}(g_i > 0) = \frac{1}{2} + \frac{1}{2} \operatorname{erf}\left(\frac{m_i}{\sqrt{2}\sigma_i}\right).
\end{equation}
A derivation of Equation~\ref{eq:rho-from-normal} is given in Appendix~\ref{sec:rho-from-normal-derivation} below.
Interestingly, this only depends on the ratio $z_i = \frac{m_i}{\sigma_i}$, and with this notation, the uncertainty-aware sign descent direction, $2\rho_i - 1$, can be expressed as
\begin{equation}
2\rho_i - 1 = \operatorname{erf}\left(\frac{z_i}{\sqrt{2}}\right).
\end{equation}

Now, a central assumption to uncertainty-aware interpretation of Adam is the following approximation to the $\operatorname{erf}$ function. 

\begin{equation}
  \label{eq:erf-approximation}
\operatorname{erf}\left(\frac{z_i}{\sqrt{2}}\right) \approx \frac{z_i}{\sqrt{1 + z_i^2}}.
\end{equation}
 A comparison plot for these two functions is given in Figure~\ref{fig:erf-approximation} in Appendix~\ref{sec:erf-approximation} below.

Also, we can multiply the numerator and denominator of the approximation by $\sigma_i$, which changes the expression to
\begin{equation}
  \label{eq:adam-update-as-z-score}
\frac{z_i}{\sqrt{1 + z_i^2}} = \frac{m_i}{\sqrt{m_i^2 + \sigma_i^2}}.  
\end{equation}
Which starts to resemble the Adam update. If we expand the denominator to see that using the definition of expectation and variance we have

\begin{equation}
  m_i^2 + \sigma_i^2 = \mathbb{E}(g_i)^2 + \big(\mathbb{E}(g_i^2) - \mathbb{E}(g_i)^2 \big) = \mathbb{E}(g_i^2)
\end{equation}
So if we replace the average of the gradient $m_i$ by the \EMA{} of the stochastic gradient, and $\sigma_i^2$ by the \EMA{} of the squared stochastic gradient, we get very close to the Adam update. 

\begin{equation}
2\rho_i - 1 \approx \frac{\mathbb{E}(g_i)}{\sqrt{\mathbb{E}(g_i^2)}} \approx \frac{\EMA_{\beta_1}[g_i]}{\sqrt{\EMA_{\beta_2}[g_i^2]}}
\end{equation}

The justification for the latter approximation is the topic of the next section.

\subsection{Derivation of Equation~\ref{eq:rho-from-normal}}
\label{sec:rho-from-normal-derivation}

Let
\begin{equation}
Z = \frac{g_i-m_i}{\sigma_i} \sim \mathcal{N}(0,1),
\end{equation}
and denote the cumulative distribution function of a standard normal random variable by $\Phi$. Since $\sigma_i>0$, standardizing $g_i$ gives
\begin{align}
\rho_i
&= \mathbb{P}(g_i>0) \\
&= \mathbb{P}\!\left(\frac{g_i-m_i}{\sigma_i}>-\frac{m_i}{\sigma_i}\right) \\
&= 1-\Phi\!\left(-\frac{m_i}{\sigma_i}\right) \\
&= \Phi\!\left(\frac{m_i}{\sigma_i}\right),
\end{align}
where the last equality follows from the symmetry of the standard normal distribution. Finally, using
\begin{equation}
\Phi(x) = \frac{1}{2}\left[1+\operatorname{erf}\!\left(\frac{x}{\sqrt{2}}\right)\right],
\end{equation}
we obtain
\begin{equation}
\rho_i
= \frac{1}{2} + \frac{1}{2}\operatorname{erf}\!\left(\frac{m_i}{\sqrt{2}\sigma_i}\right),
\end{equation}
which is Equation~\ref{eq:rho-from-normal}.

\subsection{Error-function Approximation}
\label{sec:erf-approximation}

\begin{figure}[ht]
\centering
\includegraphics[width=0.5\textwidth]{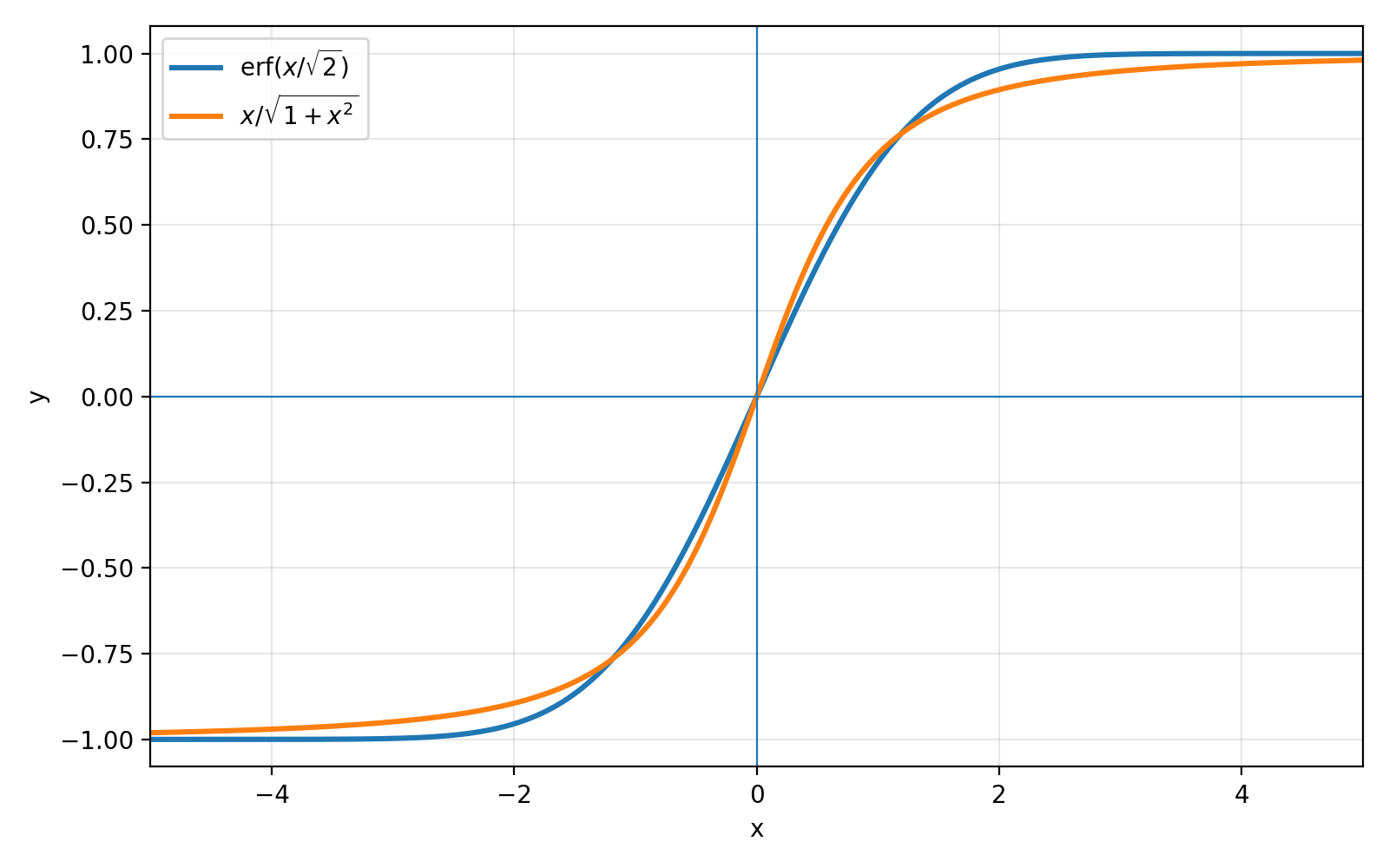}
\caption{Comparison of the error function and its approximation.}
\label{fig:erf-approximation}
\end{figure}

Figure~\ref{fig:erf-approximation} compares the error function with its rational approximation. In \citet{balles_dissecting_2020}, a similar plot is shown but is expressed in terms of $z_i^{-1}$, and is only shown when $z_i > 0$.

\subsection{Adam's \EMA{} as variational inference}
\label{sec:adam-estimates-sign-probabilities}
\citet{orvieto_search_2025} give a variational inference interpretation of the \EMA{} used in Adam optimizer. First, they show empirically that setting $\beta_1 = \beta_2$ recovers most of the performance gap between Adam and Signum. Then, they formulate a variational inference problem

\begin{equation}
m_{t+1,i},\sigma_{t+1,i}^2 = \argmin{m, \sigma^2} \Big[-\log p(g_{t+1,i}\mid m, \sigma^2) + \frac{1}{\lambda} \operatorname{KL}\big(\mathcal{N}(m_{t,i}, \sigma_{t,i}^2) || \mathcal{N}(m, \sigma^2)\big)\Big]
\end{equation}
and show that the result of plugging in the solution of this optimization problem into the equation \ref{eq:adam-update-as-z-score} gives the same formulation of adam update, namely $\EMA_{\beta}[g_i]/\sqrt{\EMA_{\beta}[g_i^2]}$. 

This last formulation has a simple interpretation, the parameterization of the Adam update tries to fit the mean and variance of the gradient distribution to the current batch gradient, while also keeping the distribution close to the previous one. The KL divergence term acts as a smoothness penalty, and the $\lambda$ parameter controls how much weight is given to the previous distribution versus the current batch gradient. This regularized variational inference approach is a design choice that we are going to use. 

Note that, in Adam, first the mean and variance of the stochastic batch gradient are estimated using smoothed out variational inference, and then the signal-to-noise ratio $z_i$ and the update direction $2\rho_i - 1$ are computed using Gaussian assumption and the approximation to the $\operatorname{erf}$ function. This ordering of operations is not necessary, and we can reverse the order of operations, and first compute $z_i$ and $\rho_i$ from the batch statistics, and then use variational inference to smooth out the estimated $\rho_i$ over time. This is the main idea behind our new optimizer, \ZEMA{}. By this change of ordering of operations, we can reduce the number of state variables per parameter from two to one, while maintaining the variance adaptation properties of Adam.

\section{Additional proofs and mathematical details}

In this section we collect the proofs and mathematical details specific to \ZEMA{}.

\subsection{Proof of Equation~\ref{eq:rho-ema}}
\label{sec:rho-ema-proof}

The problem in Equation~\ref{eq:variational-inference-for-rho} is separable across coordinates, so
fix a coordinate $i$ and a step $t$, and abbreviate the two fixed probabilities and the optimization
variable by
$$
p = \rho_{t,i}^{\text{METHOD}},
\qquad
q = \hat{\rho}_{t-1,i},
\qquad
r = \hat{\rho}_i \in (0,1).
$$
Writing out the two Bernoulli divergences, the objective is
\begin{align}
J(r)
&= \operatorname{KL}\big(\text{Bern}(p) || \text{Bern}(r)\big)
 + \lambda \operatorname{KL}\big(\text{Bern}(q) || \text{Bern}(r)\big) \\
&= p\log\frac{p}{r} + (1-p)\log\frac{1-p}{1-r}
 + \lambda\left[q\log\frac{q}{r} + (1-q)\log\frac{1-q}{1-r}\right].
\end{align}
The terms $p\log p + (1-p)\log(1-p)$ and $\lambda\big[q\log q + (1-q)\log(1-q)\big]$ do not involve
$r$; collecting them into a constant $C$ and grouping the rest by $\log r$ and $\log(1-r)$ gives
\begin{equation}
  \label{eq:rho-ema-objective}
J(r) = -\big(p + \lambda q\big)\log r
       - \big((1-p) + \lambda(1-q)\big)\log(1-r)
       + C.
\end{equation}
Note that this is where the direction of the two divergences matters: because the optimization
variable appears as the \emph{second} argument of both, it enters only through $\log r$ and
$\log(1-r)$, and the two fixed probabilities enter only through the coefficients.

Differentiating Equation~\ref{eq:rho-ema-objective},
\begin{equation}
J'(r) = -\frac{p+\lambda q}{r} + \frac{(1-p) + \lambda(1-q)}{1-r}.
\end{equation}
Setting $J'(r)=0$ and multiplying by $r(1-r)>0$,
\begin{align}
\big(p + \lambda q\big)(1-r) &= \big((1-p) + \lambda(1-q)\big)\, r, \\
p + \lambda q &= r\Big[\big(p+\lambda q\big) + \big((1-p)+\lambda(1-q)\big)\Big] = r\,(1+\lambda),
\end{align}
where the last equality uses that the two bracketed coefficients sum to $1+\lambda$. Hence the
unique stationary point is
\begin{equation}
  \label{eq:rho-ema-stationary}
r^\star = \frac{p + \lambda q}{1 + \lambda}.
\end{equation}
It is a minimum, and the only one: the second derivative
\begin{equation}
J''(r) = \frac{p+\lambda q}{r^{2}} + \frac{(1-p)+\lambda(1-q)}{(1-r)^{2}}
\end{equation}
is strictly positive for every $r \in (0,1)$, since both coefficients are nonnegative and sum to
$1+\lambda > 0$, so $J$ is strictly convex on $(0,1)$.

Finally, substituting $\beta = \frac{\lambda}{1+\lambda}$, so that $1 - \beta = \frac{1}{1+\lambda}$,
Equation~\ref{eq:rho-ema-stationary} reads
\begin{equation}
\hat{\rho}_{t,i} = \beta\,\hat{\rho}_{t-1,i} + (1-\beta)\,\rho_{t,i}^{\text{METHOD}},
\end{equation}
which applied coordinate-wise is Equation~\ref{eq:rho-ema}. As $\lambda$ ranges over $(0,\infty)$ the
smoothing parameter $\beta = \lambda/(1+\lambda)$ ranges over $(0,1)$, so every smoothing level used
in Algorithm~\ref{alg:zema} corresponds to some penalty weight $\lambda$.

\jmlrQED

\paragraph{The minimizer is interior.} Equation~\ref{eq:rho-ema-stationary} is a convex combination of
$p$ and $q$, so $r^\star \in (0,1)$ whenever $p, q \in (0,1)$, and the argument above applies as
stated. Both conditions hold in \ZEMA{}: the z-score is clipped to $\pm z_{\max}$ before either link
function is applied (Algorithm~\ref{alg:zema}, line~\ref{line:zema-z-score}), so
$\rho_{t}^{\text{CLT}}$ and $\rho_{t}^{\text{Adam}}$ are bounded away from $0$ and $1$, and the state
is initialized at $\hat{\boldsymbol{u}}_0 = \mathbf{0}$, that is $\hat{\boldsymbol\rho}_0 = \tfrac{1}{2}$.

\paragraph{Why the update is an average of probabilities.} Had both divergences been taken in the
opposite direction, with the fitted distribution as the first argument, the same calculation would
have given a weighted average of the \emph{log-odds} rather than of the probabilities,
\begin{equation}
\log\frac{r^\star}{1-r^\star}
= (1-\beta)\log\frac{p}{1-p} + \beta \log\frac{q}{1-q},
\end{equation}
with the same $\beta$. The direction chosen in
Equation~\ref{eq:variational-inference-for-rho} is therefore what makes the smoothing a plain
exponential moving average in probability space.

\paragraph{Smoothing $\boldsymbol{\rho}$ and smoothing $\boldsymbol{u}$ agree.} Because
$\boldsymbol{u} = 2\boldsymbol\rho - 1$ is affine in $\boldsymbol\rho$, applying it to
Equation~\ref{eq:rho-ema} gives
\begin{equation}
\hat{\boldsymbol{u}}_{t}
= 2\hat{\boldsymbol\rho}_{t} - 1
= \beta\big(2\hat{\boldsymbol\rho}_{t-1}-1\big) + (1-\beta)\big(2\boldsymbol\rho_{t}-1\big)
= \beta\,\hat{\boldsymbol{u}}_{t-1} + (1-\beta)\,\boldsymbol{u}_{t},
\end{equation}
the same recursion with the same $\beta$. This justifies computing the batch-level update direction
first and smoothing it afterwards, which is what Algorithm~\ref{alg:zema} does.

\subsection{Two properties of the confidence-scaled step $u_t$ in ZEMA}
\label{app:zema-theory}

Fix a coordinate and drop the index $i$. The per-example gradients $g^{(1)},\dots,g^{(B)}$ are i.i.d.\ with mean $\mu$ and finite variance $s^2>0$; $m$ is their mean, $\sigma = s/\sqrt{B}$ its standard error, $z = m/\sigma$, and $u = h(z)$ with $h(z)=\operatorname{erf}(z/\sqrt2)$. Write $\zeta = \sqrt{B}\,\mu/s$ for the population evidence. All statements hold for any link $h$ that is continuous, odd, strictly increasing and bounded by $1$ with $h(z)\to\pm1$ as $z\to\pm\infty$, which includes the Adam-shaped link $z(1+z^2)^{-1/2}$; the closed forms are for $\operatorname{erf}$.

\begin{proposition}[Saturation at large batch]
\label{prop:saturation}
If $\mu\neq 0$, then $u\to\operatorname{sign}(\mu)$ almost surely as $B\to\infty$. The same holds when $\sigma$ is replaced by any consistent estimate $\hat\sigma$.
\end{proposition}

\begin{proof}
By the strong law of large numbers $m\to\mu$ a.s., so $z=\sqrt{B}\,m/s\to\operatorname{sign}(\mu)\cdot\infty$ a.s. Since $h$ is continuous with $h(z)\to\pm1$, $h(z)\to\operatorname{sign}(\mu)$ a.s. With $\hat\sigma = \hat s/\sqrt{B}$ and $\hat s\to s$ a.s., the ratio $m/\hat\sigma$ has the same limit.
\end{proof}

Hence for every coordinate with a non-zero full-dataset gradient, the \ZEMA{} step converges to the Signum step, and the damping in Proposition~\ref{prop:damping} disappears. Adam's step $m/\sqrt{v}$ does not share this limit: $v$ is a temporal average of squared batch gradients, so $v \approx \bar\mu^2 + \operatorname{Var}_t(\mu_t) + s^2/B$ where $\operatorname{Var}_t(\mu_t)$ is the variation of the mean gradient across steps. As $B\to\infty$ the last term vanishes but the middle one does not, so $|m/\sqrt v|$ stays strictly below one whenever the gradient moves during training.

\begin{proposition}[Damping at weak evidence]
\label{prop:damping}
Assume the per-example gradients are Gaussian, so that $z\sim\mathcal N(\zeta,1)$. Then
\begin{enumerate}
\item $|u|<1$ for every finite $z$, whereas the Signum step satisfies $|u|=1$;
\item $\mathbb E[u] = \operatorname{erf}(\zeta/2)$, which is odd, strictly increasing in $\zeta$, has the sign of $\mu$, and vanishes at $\zeta=0$;
\item at $\zeta = 0$, $\mathbb E[u^2] = \tfrac13$.
\end{enumerate}
\end{proposition}

\begin{proof}
(1) is immediate from $|\operatorname{erf}(\cdot)|<1$ on $\mathbb R$. For (2), write $u = 2\Phi(z)-1$. For $z\sim\mathcal N(\zeta,1)$ and an independent $W\sim\mathcal N(0,1)$, $\mathbb E[\Phi(z)] = \mathbb P(W\le z) = \mathbb P(W - z\le 0) = \Phi(\zeta/\sqrt2)$, since $W-z\sim\mathcal N(-\zeta,2)$. Hence $\mathbb E[u] = 2\Phi(\zeta/\sqrt2)-1 = \operatorname{erf}(\zeta/2)$; the listed properties are those of $\operatorname{erf}$. For (3), at $\zeta=0$ the variable $\Phi(z)$ is uniform on $(0,1)$, so $\mathbb E[(2\Phi(z)-1)^2] = \mathbb E[(2U-1)^2] = \tfrac13$.
\end{proof}

Property (2) says the expected per-coordinate step is $\eta\,\operatorname{erf}(\zeta/2)$: the learning rate is scaled down exactly where the batch gives weak evidence, and the scaling is a fixed function of the evidence rather than of the gradient magnitude. Property (3) quantifies the damping at a stationary coordinate: \ZEMA{}'s mean squared step there is one third of Signum's, which is the mechanism we associate with the reduced seed-to-seed variance in Tables~\ref{tab:finetuning} and~\ref{tab:pretraining}. Monotonicity in (2) holds without the Gaussian assumption for any symmetric noise, since $z$ then belongs to a location family and $h$ is increasing; the closed form does not. When $\sigma$ is estimated from $k$ examples, $z$ is Student-$t$ with $k-1$ degrees of freedom rather than Gaussian; replacing $\Phi$ by the corresponding $t$ distribution gives the \ZEMA{}-$t$ which we have not showed empirically since it performs approximately the same as \ZEMA{}-CLT.

\clearpage
\section{Experimental details and additional experiments}
\label{app:additional-experiments}

In this section we report the fine-tuning, instruction fine-tuning and
pretraining details, and add a number of additional experiments to further
validate \ZEMA{}.

\subsection{E2E and SAMSum fine-tuning details}
\label{sec:finetuning-details}

\paragraph{Setup.} LoRA adapters of rank $16$ with $\alpha=32$ are attached to
seven projections per block; the base weights are frozen in bfloat16 and the
adapters held in float32. E2E NLG is trained for five epochs at batch $64$ with a
linear schedule ($3{,}290$ steps); SAMSum for two epochs at batch $16$ with a
cosine schedule and $6\%$ warmup ($1{,}842$ steps). The batch is held fixed
across all four models, because the \ZEMA{} statistic is a $z$-score of a
$B$-sample mean and changing $B$ between cells would change the estimator rather
than only the model. This is also why the Qwen3-1.7B numbers here supersede any
run of ours at a different batch: at a different $B$ the arm is not the same
estimator, so such runs cannot be placed on the same axis. All arms share $\beta=0.95$ (the \ZEMA{} smoothing and AdamW's
$\beta_1=\beta_2$), $\sigma_0=10^{-8}$, and $z_{\max}=10$. Runs are on a single
NVIDIA L40S.

\paragraph{Selection metric.} We select on the \emph{minimum} validation
perplexity over epochs, not the final-epoch value. On E2E every cell turns over
between epochs two and four, so the final-epoch value measures where a run
happened to sit on its own overfitting curve rather than the quality of the rate;
ranking on it bends the sweep curves into a shape that is an artifact of the
epoch budget. On SAMSum, which trains for two epochs and never turns over, the
minimum and the final value coincide in every cell, so the choice changes nothing
there.

\paragraph{Rate selection.} Each optimizer is swept independently on each
(model, task) cell, at a single seed, with generation metrics disabled. On the
six Llama cells the initial grid is $\{2, 5\}\cdot10^{-4}$ and
$\{1, 2\}\cdot10^{-3}$; it was then extended downwards---through $10^{-4}$,
$5\cdot10^{-5}$, $2.5\cdot10^{-5}$, $10^{-5}$ and $5\cdot10^{-6}$---for whichever
cells needed it, until every cell's argmin had \emph{two} finished grid points on
each side. We adopted this counting rule rather than a significance test on the
neighbouring point, because the question a bracket has to answer is whether
anything below the current edge is better, not whether the edge beats its
neighbour by more than noise. The two Qwen cells were swept in a single pass on
the wider grid those extensions had established,
$\{5\cdot10^{-5}, 10^{-4}, 2\cdot10^{-4}, 5\cdot10^{-4}, 10^{-3}, 2\cdot10^{-3}\}$;
every argmin is interior, and all but one have two points on each side. The
exception is \ZEMA{}-Adam on E2E, whose argmin of $10^{-3}$ has two points below and
only one above; we report it as selected because the point above it is worse by
$0.058$ perplexity, roughly ten times the seed standard deviation of that cell,
but it is a weaker bracket than every other cell in this table.

The selected rate is then frozen and rerun with seeds $2$ and $3$; it is not
re-derived at report time, so a later extension of the grid cannot silently
repoint a seed that has already run.

\begin{figure}[t]
\centering
\includegraphics[width=\textwidth]{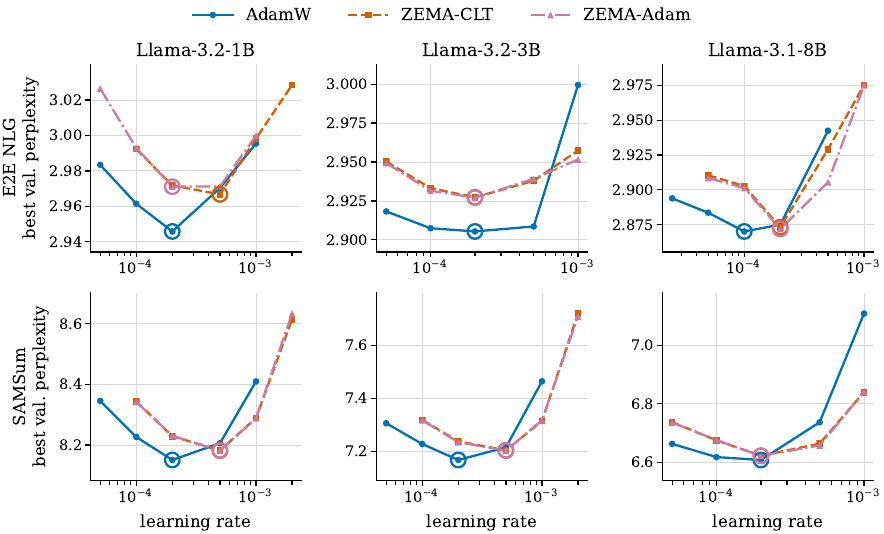}
\caption{Learning-rate sweep, single seed. Each panel is one (model, task) cell;
the marker is the best validation perplexity over epochs at that rate and the
ring marks the selected rate. Each arm is shown as its optimum plus the two
grid points on either side; grids were extended downwards per arm until that was
possible, so the arms do not all span the same range. Note the vertical scales
differ between panels.}
\label{fig:lr-sweep}
\end{figure}

Figure~\ref{fig:lr-sweep} shows the resulting sweeps and
Table~\ref{tab:selected-rates} the rate each cell selected. The perplexities in
the figure are single-seed, from the sweep stage, and are therefore not directly
comparable to the three-seed means in Table~\ref{tab:finetuning}; they establish
that every optimum is interior and give the scale of one grid step. Signum was also swept on every cell but is omitted from the tables: on Llama its
optimum sits an order of magnitude below every other arm and it loses at every
rung, so it is not part of this comparison. We note in passing that on Qwen3-1.7B
it does not lose---its paired differences from AdamW are $+0.002 \pm 0.032$ on
E2E and $+0.003 \pm 0.006$ on SAMSum, both indistinguishable---at an optimum of
$10^{-4}$, five times below AdamW's rather than ten. We have not investigated
this and do not draw a conclusion from a single model.

Two features of Figure~\ref{fig:lr-sweep} are worth noting. The curves are
locally flat near their minima, so on most cells one grid step costs between
$0.002$ and $0.02$ perplexity---of the same order as, and usually larger than,
the entire gap between AdamW and \ZEMA{} in Table~\ref{tab:finetuning}. And the two
\ZEMA{} variants track each other almost exactly at every rate, which is expected:
their links, $\operatorname{erf}(z/\sqrt{2})$ and $z(1+z^{2})^{-1/2}$, agree
closely over the range of $z$ these runs occupy (Appendix~\ref{sec:erf-approximation}).

\begin{figure}[t]
\centering
\includegraphics[width=0.86\textwidth]{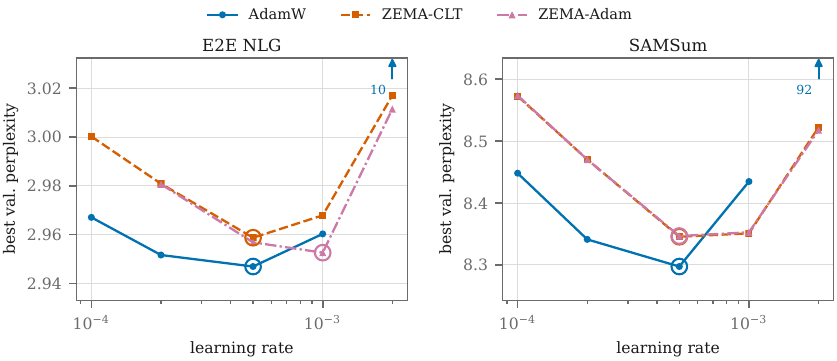}
\caption{Qwen3-1.7B learning-rate sweep, single seed, plotted on the same
conventions as Figure~\ref{fig:lr-sweep}: the marker is the best validation
perplexity over epochs and the ring marks the selected rate. AdamW diverges at
$2\cdot10^{-3}$ on both tasks; those two points are drawn as arrows with their
values rather than clipped, since they set the top of the grid. Note the
vertical scales differ between panels and from Figure~\ref{fig:lr-sweep}.}
\label{fig:qwen-lr-sweep}
\end{figure}

\begin{table}[t]
\centering
\caption{Selected learning rate per cell, from the sweeps in
Figures~\ref{fig:lr-sweep} and~\ref{fig:qwen-lr-sweep}. These rates are frozen and reused for the three-seed
runs in Table~\ref{tab:finetuning}.}
\label{tab:selected-rates}
\small
\begin{tabular}{llccc}
\toprule
Model & Task & AdamW & \ZEMA{}-CLT & \ZEMA{}-Adam \\
\midrule
\multirow{2}{*}{Llama-3.2-1B} & E2E & $2{\cdot}10^{-4}$ & $5{\cdot}10^{-4}$ & $2{\cdot}10^{-4}$ \\
 & SAMSum & $2{\cdot}10^{-4}$ & $5{\cdot}10^{-4}$ & $5{\cdot}10^{-4}$ \\
\midrule
\multirow{2}{*}{Llama-3.2-3B} & E2E & $2{\cdot}10^{-4}$ & $2{\cdot}10^{-4}$ & $2{\cdot}10^{-4}$ \\
 & SAMSum & $2{\cdot}10^{-4}$ & $5{\cdot}10^{-4}$ & $5{\cdot}10^{-4}$ \\
\midrule
\multirow{2}{*}{Llama-3.1-8B} & E2E & $1{\cdot}10^{-4}$ & $2{\cdot}10^{-4}$ & $2{\cdot}10^{-4}$ \\
 & SAMSum & $2{\cdot}10^{-4}$ & $2{\cdot}10^{-4}$ & $2{\cdot}10^{-4}$ \\
\midrule
\multirow{2}{*}{Qwen3-1.7B} & E2E & $5{\cdot}10^{-4}$ & $5{\cdot}10^{-4}$ & $1{\cdot}10^{-3}$ \\
 & SAMSum & $5{\cdot}10^{-4}$ & $5{\cdot}10^{-4}$ & $5{\cdot}10^{-4}$ \\
\bottomrule
\end{tabular}
\end{table}

\clearpage
\subsection{SlimPajama details}
\label{sec:pretraining-details}

\paragraph{Setup.} A decoder-only transformer with $12$ layers,
$d_{\text{model}}=768$, $12$ heads, SwiGLU feed-forward of width $2048$, RoPE,
context $2048$ and an \emph{untied} embedding and output head over the
GPT-NeoX $50{,}280$-token vocabulary: $162{,}183{,}936$ parameters, matching the
configuration of \citet{orvieto_search_2025}. Global batch $256$ sequences
($8$ micro-batches of $8$ on each of four ranks), $6{,}103$ steps, $3.2$B
tokens. Linear warmup over $610$ steps then cosine decay to $10^{-5}$; weight
decay $0.1$ on matrices only, gradient clipping $1.0$, and $\beta=0.95$ for
every arm---AdamW in the equal-$\beta_1=\beta_2$ form that paper identifies as
the strong baseline. Four NVIDIA L40S.

As in the fine-tuning campaign the \ZEMA{} arms are hybrid: the estimator covers
the body linears ($84.9$M parameters) while the embedding, the untied head and
the norm gains stay on AdamW at the same rate, leaving each arm a single
learning-rate knob.

\paragraph{Corpus.} The paper's \texttt{cerebras/\allowbreak SlimPajama-627B} no longer
resolves, so we use instead \texttt{DKYoon/\allowbreak SlimPajama-6B}, a published subset of the
same source corpus, with its own held-out validation split ($9.2$M tokens)
rather than a resampled tail of the training data.

\paragraph{What was and was not tuned.} AdamW at $8\cdot10^{-3}$ and Signum at
$10^{-3}$ are the rates \citet{orvieto_search_2025} select at this scale; we ran
no search of our own. The \ZEMA{} arms have no published rate to copy, and rather
than sweep them---which would have given them an advantage the baselines did not
get---both were run at AdamW's rate. The result in
Table~\ref{tab:pretraining} is therefore a lower bound on what \ZEMA{} does here.

\paragraph{Distance from the published number.} Our AdamW reaches $22.79 \pm
0.46$ against the $21.86$ that paper reports, so this setup runs somewhat harder
than theirs---plausibly the corpus substitution above. That offset applies to
every arm equally and does not affect the within-table comparison, which is why
Table~\ref{tab:pretraining} is reported as paired differences against our own
AdamW rather than against the published value.

\paragraph{Distributed correctness.} All four arms trained on four L40S under
DDP. The \ZEMA{} arms do not use \texttt{DistributedDataParallel}: the bound
weights carry no \texttt{.grad}, and their evidence statistic is pooled across
ranks by a hand-written all-reduce, verified against a single-GPU run to
$4\cdot10^{-8}$ relative error in fp32 before the campaign was launched.

\begin{figure}[h]
\centering
\includegraphics[width=\textwidth]{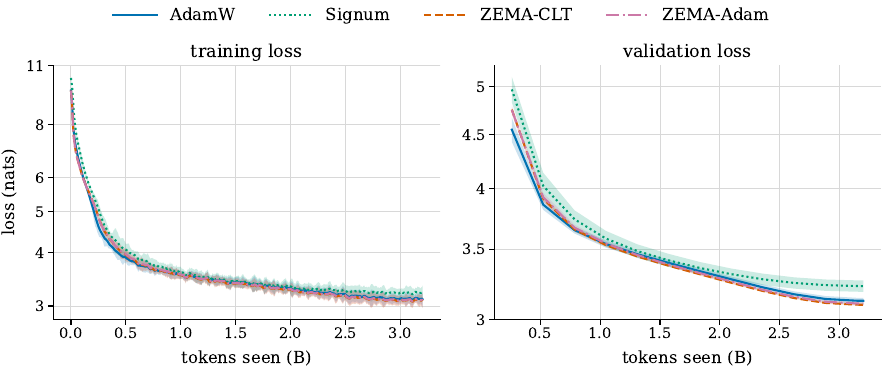}
\caption{Training and validation loss over the replication run, mean of three
seeds, on a logarithmic vertical scale. The shaded band is the $95\%$
Student-$t$ interval on the seed mean, the same interval as
Table~\ref{tab:pretraining}. Training loss is logged every $10$ steps and is
shown under a centred $50$-step rolling mean, applied identically to the mean
and to both band edges; validation is evaluated every $500$ steps and is
unsmoothed.}
\label{fig:replication-curves}
\end{figure}

\paragraph{Curves.} Figure~\ref{fig:replication-curves} shows the full
trajectories. Signum separates from the other three within the first $0.5$B
tokens and never rejoins them; AdamW and the two \ZEMA{} arms stay together until
roughly $1.5$B, after which the \ZEMA{} curves sit marginally below. The seed bands
are widest early and narrow as the runs converge---at the end the \ZEMA{}-CLT band
is about a fifth the width of AdamW's, the same ordering of seed-to-seed spread
that Table~\ref{tab:pretraining} reports.

\paragraph{Paired differences over the run.}
Table~\ref{tab:slim-horizon} tabulates what Figure~\ref{fig:replication-curves}
plots, as paired differences at matched seed. The run has the same three phases
as the TinyStories campaign of Appendix~\ref{app:horizon}: AdamW leads by
$21.45$ perplexity at the first checkpoint, the arms cross at about $1.1$B
tokens, and from $1.31$B onward all three seeds favour \ZEMA{}-CLT at every
checkpoint. The difference then settles at $-0.6$ perplexity and stays there,
resolving at the $95\%$ level over the final four checkpoints. Unlike
TinyStories, where the gap closes as the cosine anneals, here it persists to the
endpoint of Table~\ref{tab:pretraining}.

\begin{table}[h]
\centering
\caption{SlimPajama, validation perplexity against token budget, three seeds
per cell. $\Delta$ is \ZEMA{}-CLT minus AdamW at matched seed, with a two-sided
$95\%$ $t$-interval over the three paired differences; negative favours
\ZEMA{}. The final column counts seeds favouring \ZEMA{}. Bold marks intervals
excluding zero. \ZEMA{}-t2 is shown for reference and does not enter the paired
test. The final row is the endpoint reported in
Table~\ref{tab:pretraining}.}
\label{tab:slim-horizon}
\small
\begin{tabular}{lrrrrrc}
\toprule
Tokens & AdamW & \ZEMA{}-CLT & \ZEMA{}-t2 & $\Delta$ & $95\%$ CI & seeds \\
\midrule
$0.26$B & $94.76$ & $116.22$ & $115.84$ & $\mathbf{+21.45}$ & $[+8.18, +34.73]$ & 0/3 \\
$0.52$B & $47.49$ & $49.94$  & $50.57$  & $+2.45$  & $[-0.60, +5.50]$ & 0/3 \\
$0.79$B & $38.49$ & $38.73$  & $39.24$  & $+0.24$  & $[-1.46, +1.94]$ & 1/3 \\
$1.05$B & $34.50$ & $34.20$  & $34.64$  & $-0.29$  & $[-1.64, +1.06]$ & 1/3 \\
$1.31$B & $31.90$ & $31.37$  & $31.70$  & $-0.52$  & $[-1.50, +0.46]$ & 3/3 \\
$1.57$B & $29.89$ & $29.26$  & $29.52$  & $-0.62$  & $[-1.54, +0.30]$ & 3/3 \\
$1.83$B & $28.09$ & $27.48$  & $27.71$  & $-0.61$  & $[-1.44, +0.22]$ & 3/3 \\
$2.10$B & $26.49$ & $25.85$  & $26.07$  & $-0.64$  & $[-1.36, +0.08]$ & 3/3 \\
$2.36$B & $25.02$ & $24.40$  & $24.59$  & $-0.63$  & $[-1.29, +0.03]$ & 3/3 \\
$2.62$B & $23.84$ & $23.22$  & $23.41$  & $\mathbf{-0.62}$ & $[-1.21, -0.03]$ & 3/3 \\
$2.88$B & $23.07$ & $22.45$  & $22.64$  & $\mathbf{-0.62}$ & $[-1.18, -0.05]$ & 3/3 \\
$3.15$B & $22.80$ & $22.19$  & $22.37$  & $\mathbf{-0.61}$ & $[-1.16, -0.06]$ & 3/3 \\
$3.20$B & $22.79$ & $22.18$  & $22.36$  & $\mathbf{-0.61}$ & $[-1.16, -0.06]$ & 3/3 \\
\bottomrule
\end{tabular}
\end{table}

\clearpage
\subsubsection{SlimPajama learning-rate selection}
\label{sec:pretraining-choices}

\begin{figure}[h]
\centering
\includegraphics[width=0.82\textwidth]{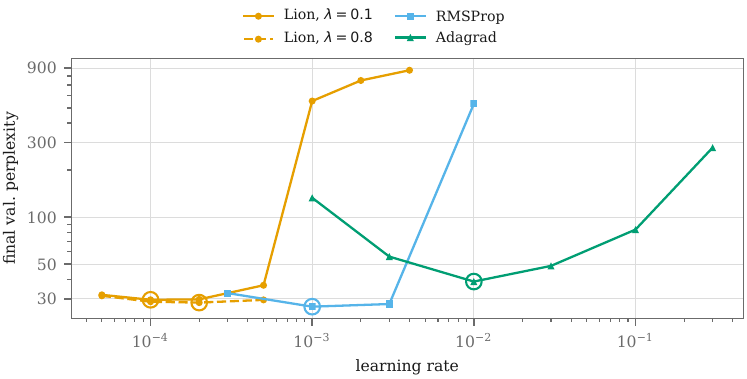}
\caption{Pretraining sweeps for the tuned baselines of
Table~\ref{tab:pretraining}; the ring marks the selected rate and the values
are those of Table~\ref{tab:pretraining-sweeps}. \emph{Both axes are
logarithmic.} A linear vertical axis is unusable here: the diverged cells reach
perplexity $871$ against optima near $30$, and the divergence is the finding
for Lion rather than something to clip away. A weight decay is a knob on an
optimizer and not a different optimizer, so Lion's two decays share a colour
and are separated by the dash pattern. Note that the arms operate at genuinely
different rate scales---Lion's optimum is $80\times$ below AdamW's and
Adagrad's is above it---which is why the horizontal span is so wide.}
\label{fig:pretrain-sweep-baselines}
\end{figure}

Every rate in Table~\ref{tab:pretraining} is either taken from
\citet{orvieto_search_2025} or is the argmin of a sweep at seed $1$ over the
grid below. We report the whole grid, not only the winner: a rate quoted
without its neighbours cannot be checked, and a selection procedure that stops
at the first interior-looking point is exactly how a sweep reports a boundary as
an optimum. Every argmin below is interior to its own grid.
Table~\ref{tab:pretraining-sweeps} lists the values and
Figure~\ref{fig:pretrain-sweep-baselines} plots them.

\begin{table}[h]
\centering
\caption{Pretraining learning-rate sweeps. Validation perplexity at seed $1$;
$\ast$ marks the selected rate, which is then rerun with seeds $2$ and $3$.
AdamW and Signum are not swept---they use the rates \citet{orvieto_search_2025}
reports---and the \ZEMA{} arms deliberately copy AdamW's rate, so that the arm is
the only thing that differs from the baseline.}
\label{tab:pretraining-sweeps}
\small
\begin{tabular}{ll}
\toprule
Arm & validation perplexity by learning rate (seed $1$) \\
\midrule
Lion, $\lambda=0.1$ & $\underset{5\cdot10^{-5}}{31.8}$\ \ $\underset{10^{-4}}{\ast 29.6}$\ \ $\underset{2\cdot10^{-4}}{29.8}$\ \ $\underset{5\cdot10^{-4}}{36.7}$\ \ $\underset{10^{-3}}{554}$\ \ $\underset{2\cdot10^{-3}}{749}$\ \ $\underset{4\cdot10^{-3}}{871}$ \\[4pt]
Lion, $\lambda=0.8$ & $\underset{5\cdot10^{-5}}{31.4}$\ \ $\underset{10^{-4}}{28.8}$\ \ $\underset{2\cdot10^{-4}}{\ast 28.4}$\ \ $\underset{5\cdot10^{-4}}{29.5}$ \\[4pt]
RMSProp & $\underset{3\cdot10^{-4}}{32.7}$\ \ $\underset{10^{-3}}{\ast 26.7}$\ \ $\underset{3\cdot10^{-3}}{27.8}$\ \ $\underset{10^{-2}}{531}$ \\[4pt]
Adagrad & $\underset{10^{-3}}{133}$\ \ $\underset{3\cdot10^{-3}}{55.9}$\ \ $\underset{10^{-2}}{\ast 38.7}$\ \ $\underset{3\cdot10^{-2}}{48.8}$\ \ $\underset{10^{-1}}{83.3}$\ \ $\underset{3\cdot10^{-1}}{279}$ \\
\bottomrule
\end{tabular}
\end{table}

\paragraph{Lion's grid is an order of magnitude below its published one.} Lion's
recommended rate for language-model pretraining is $3$--$10\times$ smaller than
AdamW's, which at AdamW's $8\cdot10^{-3}$ would put it near $10^{-3}$. Every
rate in that band diverged here: $10^{-3}$, $2\cdot10^{-3}$ and $4\cdot10^{-3}$
finish at perplexity $554$, $749$ and $871$. The grid had to be moved a full
decade down before Lion trained at all, and its optimum sits at $10^{-4}$,
$80\times$ below AdamW's. We report this because a reader reproducing Lion from
its published guidance at this scale will see divergence, not a weak result.

\paragraph{Weight decay: $\lambda=0.1$ is reported, $\lambda=0.8$ is a control.}
Lion is normally run with a decoupled decay several times larger than AdamW's,
because its update is a sign and so has a fixed magnitude that does not shrink
as gradients do. Table~\ref{tab:pretraining} reports $\lambda=0.1$, the value
every other arm uses, so that the table varies one thing. We also ran Lion's own
$\lambda=0.8$ with its own sweep: it selects $2\cdot10^{-4}$ and reaches
$27.97 \pm 1.53$, a paired improvement of $1.13 \pm 0.24$ over the matched-decay
row. So the larger decay does help, and materially---but it does not change the
ordering in Table~\ref{tab:pretraining}, where Lion trails AdamW by $6.32$;
$\lambda=0.8$ would still trail by $5.19$. Promoting the $\lambda=0.8$ row into
the main table would silently put an $8\times$ regularization difference inside
a comparison whose point is that only the optimizer changes.

\paragraph{Lion with $\beta_1=\beta_2$ is Signum, exactly.} This is an algebraic
identity, not an approximation, and it is worth stating because it constrains
what a Lion baseline can show. Lion forms its update from an interpolation and
then updates its momentum,
\[
u_t = \operatorname{sign}\!\big(\beta_1 m_{t-1} + (1-\beta_1) g_t\big),
\qquad
m_t = \beta_2 m_{t-1} + (1-\beta_2) g_t .
\]
Setting $\beta_1 = \beta_2 = \beta$ makes the two expressions the same tensor,
so $u_t = \operatorname{sign}(m_t)$ with $m_t$ the Signum momentum: the arms
coincide for every input, at every step. Running Lion at the $\beta=0.95$ this
paper's other arms share would therefore not have produced a Lion baseline at
all, only a second copy of the Signum row. Lion is run at its own published
$(\beta_1, \beta_2) = (0.9, 0.99)$ for this reason, and it is the one arm in
Table~\ref{tab:pretraining} not matched to $\beta=0.95$.

\paragraph{Adagrad has no decay to match.} Adagrad accumulates an undecayed sum
of squared gradients; adding a decay to it is what makes it RMSProp. It
therefore has no $\beta$ to set to $0.95$, and the two arms together span that
axis: RMSProp is the same second-moment normalization at $\rho=0.95$, and is
also exactly AdamW at $\beta_1=\beta_2=0.95$ with the first moment removed.
That makes the pair a clean ablation of the two AdamW moments taken separately,
which is how they should be read---not as competitive optimizers at this scale.

\paragraph{Reading the one-state baselines.} The four one-state arms of
Table~\ref{tab:pretraining} separate widely, and which moment they keep does
not predict where they land: keeping a first moment and taking its sign gives
$25.25$ (Signum) and $29.10$ (Lion), while keeping a second and normalizing by
it gives $26.71$ (RMSProp) and $38.31$ (Adagrad). Signum is the informative
arm, as the one that takes the sign without scaling it by a confidence: its
$2.46$ deficit is what dropping that scaling costs in this setting
(Appendix~\ref{sec:uncertainty-adaptation}). \ZEMA{}'s single state is of a
different kind from any of these---the smoothed confidence, formed from the
batch mean and its standard error together---which is why it is the only
one-state arm in the table that does not pay for the reduction.

\clearpage
\subsection{Instruction fine-tuning details (MATH and TULU)}
\label{app:ppl-sft}

In this section we present additional instruction fine-tuning experiments to evaluate non-stationarity. Instruction fine-tuning illustrates this because task mixtures can introduce significant drift (TULU) and rare tokens (MATH).
We fine-tune Qwen2.5-0.5B (non-instruction-tuned) on a controlled
dialogue/math/code Tulu-3 drift mixture using rank-8 LoRA adapters
($4.40$M trainable parameters, sequence length 512, $128$ validation examples
per source), and Llama-3.2-1B on the rare-token MATH split using rank-16 LoRA
adapters ($11.27$M trainable parameters, sequence length 1024, $754$ validation
examples). Each cell performs exactly $48$ optimizer updates at a global batch
drawn from $\{8, 32, 128, 512\}$, drawing examples with replacement where the
cell's example budget exceeds the dataset, and is evaluated at eight
checkpoints; we report the best validation loss over those checkpoints,
exponentiated to a token-level perplexity. Adapters are held in
\texttt{float32} with weight decay $0.01$ and $\beta = 0.95$ for both
optimizers. Both campaigns use a single selection seed ($11$), so the tables
below carry no confidence intervals and differences should be compared against
the scale of seed variation rather than treated as resolved.

\begin{table}[t]
\centering
\caption{Validation perplexity for instruction fine-tuning at fixed update
count ($48$ steps), with the independently selected learning rate for each
cell. $\Delta$ is \ZEMA{}-CLT minus AdamW; positive values favour AdamW.
$^{\dagger}$ marks a cell whose selected learning rate is the largest searched,
so its optimum is not bracketed above.}
\label{tab:ppl-sft}
\begin{tabular}{lrlrlrr}
\toprule
Batch & AdamW & lr & \ZEMA{}-CLT & lr & $\Delta$ & $\Delta\,\%$ \\
\midrule
\multicolumn{7}{l}{\emph{Tulu-3 drift, Qwen2.5-0.5B, rank-8 LoRA}} \\
$8$   & $2.4568$ & $4\!\times\!10^{-4}$ & $2.4594$ & $1\!\times\!10^{-3}$ & $+0.0026$ & $+0.11$ \\
$32$  & $2.4247$ & $2\!\times\!10^{-4}$ & $2.4266$ & $5\!\times\!10^{-4}$ & $+0.0019$ & $+0.08$ \\
$128$ & $2.3968$ & $5\!\times\!10^{-4}$ & $\mathbf{2.3915}$ & $1\!\times\!10^{-3}$ & $-0.0053$ & $-0.22$ \\
$512$ & $2.3739$ & $5\!\times\!10^{-4}$ & $\mathbf{2.3720}$ & $1\!\times\!10^{-3}$ & $-0.0020$ & $-0.08$ \\
\midrule
\multicolumn{7}{l}{\emph{MATH rare-token split, Llama-3.2-1B, rank-16 LoRA}} \\
$8$   & $2.7018$ & $2\!\times\!10^{-4}$ & $\mathbf{2.6997}$ & $1\!\times\!10^{-3}$ & $-0.0021$ & $-0.08$ \\
$32$  & $2.6364$ & $5\!\times\!10^{-4}$ & $2.6373$ & $2\!\times\!10^{-3}$ & $+0.0009$ & $+0.03$ \\
$128$ & $2.5503$ & $1\!\times\!10^{-3}$ & $2.5577$ & $2\!\times\!10^{-3}$ & $+0.0074$ & $+0.29$ \\
$512$ & $2.5116$ & $1\!\times\!10^{-3}$ & $2.5165^{\dagger}$ & $2\!\times\!10^{-3}$ & $+0.0048$ & $+0.19$ \\
\bottomrule
\end{tabular}
\end{table}

In the parameter-efficient setting the two optimizers are indistinguishable.
Every difference in Table~\ref{tab:ppl-sft} falls within $\pm 0.3\%$, and the
sign is not consistent: \ZEMA{} is nominally ahead at the two largest Tulu-3
batches and at the smallest MATH batch, behind elsewhere. With a single
selection seed these margins are smaller than the seed variation one should
expect from a $48$-update run, and we therefore read the fine-tuning results as
parity rather than as evidence for either method. Two observations survive this
caveat. First, the fixed-update budget removes the large-batch degradation seen
in the fixed-token setting: \ZEMA{} does not fall behind as the batch grows, and
at $B = 128$ and $B = 512$ on Tulu-3 it is marginally ahead, so the degradation
reported elsewhere is at least partly attributable to update count rather than
to batch size as such. Second, \ZEMA{}'s selected learning rate is consistently
two to four times AdamW's at matched batch, a systematic offset rather than
per-cell scatter; a shared grid would have handicapped \ZEMA{}, and we recommend
independent selection for any comparison of this kind. We note that the MATH
$B = 512$ \ZEMA{} cell remains unbracketed above, so its already small deficit may
narrow further.

\subsection{Sign estimation accuracy in SlimPajama pre-training}
\label{app:sign-estimation-accuracy}
\label{app:sign-slimpajama}

To isolate sign estimation from trajectory divergence, a separate probe
trains one shared AdamW trajectory. At every checkpoint, shadow AdamW and
\ZEMA{} states receive the same training batch at the same parameter vector;
neither shadow state changes the trajectory. We instrument the QKV matrix in
block zero and the untied output head. The reference direction is a
deterministic held-out gradient, disjoint from training and reused at every
checkpoint to compute the trget gradient at that checkpoint.

For direction $d$ and held-out gradient $g^\star$, \emph{accuracy} is the
coordinate mean of
$\mathbf{1}\{\operatorname{sign}(d_i)=\operatorname{sign}(g_i^\star)\}$. The
$\Delta$ columns are \ZEMA{} minus AdamW in absolute percentage points; positive
values favour \ZEMA{}.

We run the probe on a SlimPajama model (12 layers,
$d_{\mathrm{model}} = 768$, 12 heads, FFN width 2048, vocabulary $50{,}280$,
context 2048) trained at global batch $256$ for $100$M tokens, probing every
$48$ updates against a fixed $256$-sequence oracle, with three seeds.
Table~\ref{tab:sign-slimpajama} reports every probe rather than only the last,
since the two layer groups evolve differently over training.

\begin{table}[t]
\centering
\caption{Sign agreement with a fixed $256$-sequence oracle on SlimPajama,
global batch $256$, three seeds. At the first probe the two optimizers' state
is still identical by construction, so the matching accuracies there serve as a
sanity check on the setup.}
\label{tab:sign-slimpajama}
\begin{tabular}{lrrrrrr}
\toprule
& \multicolumn{3}{c}{Body QKV} & \multicolumn{3}{c}{Output head} \\
\cmidrule(lr){2-4} \cmidrule(lr){5-7}
Step & AdamW & \ZEMA{}-CLT & $\Delta$ & AdamW & \ZEMA{}-CLT & $\Delta$ \\
\midrule
$1$   & $68.30$ & $68.30$ & $+0.00 \pm 0.00$ & $69.84$ & $69.84$ & $+0.00 \pm 0.00$ \\
$48$  & $50.38$ & $55.30$ & $+4.92 \pm 7.60$ & $52.34$ & $\mathbf{61.68}$ & $\mathbf{+9.35} \pm 0.90$ \\
$96$  & $53.53$ & $54.62$ & $+1.09 \pm 3.25$ & $54.09$ & $\mathbf{63.68}$ & $\mathbf{+9.59} \pm 0.73$ \\
$144$ & $55.44$ & $51.62$ & $-3.82 \pm 5.65$ & $52.20$ & $\mathbf{63.73}$ & $\mathbf{+11.53} \pm 0.67$ \\
$190$ & $56.13$ & $53.65$ & $-2.48 \pm 6.37$ & $52.69$ & $\mathbf{62.10}$ & $\mathbf{+9.41} \pm 1.45$ \\
\bottomrule
\end{tabular}
\end{table}

The two layer groups separate sharply. The output head shows a large and
stable gain, between $+9.4$ and $+11.5$ points at every probe after the first, with
seed standard deviations below $1.5$ points; this is the clearest sign-recovery
result we obtain anywhere. Body QKV, by contrast, is not separable from zero at
any probe: the seed standard deviation exceeds the mean at all four, and the
point estimate changes sign over training, drifting from $+4.92$ at step $48$
to $-2.48$ at step $190$. We therefore report a sign-recovery advantage for
this model only on the output head, and explicitly not on the body, and we do
not claim a general mechanism.

\subsection{Sign estimation accuracy in fine-tuning}
\label{app:sign-lora}

We repeat the probe on the LoRA setting of the main text: Llama-3.1-8B-Instruct
SAMSum ($B = 16$, $1{,}842$ updates), probing every $460$ updates against a
$64$-example oracle with three seeds. Two adapter tensors are tracked, an early
attention projection and a late MLP projection.

\begin{table}[t]
\centering
\caption{Sign agreement on LoRA adapter tensors at the final probe, three
seeds. Tensor names abbreviate \texttt{lora\_B} projections by layer index.}
\label{tab:sign-lora}
\begin{tabular}{llrrr}
\toprule
Task & Tensor & AdamW & \ZEMA{}-CLT & $\Delta$ \\
\midrule
E2E    & L0 \texttt{q\_proj}      & $51.50$ & $51.29$ & $-0.21 \pm 0.43$ \\
E2E    & L31 \texttt{down\_proj}  & $51.61$ & $51.54$ & $-0.07 \pm 0.60$ \\
SAMSum & L0 \texttt{q\_proj}      & $48.28$ & $48.82$ & $+0.54 \pm 0.44$ \\
SAMSum & L31 \texttt{down\_proj}  & $52.58$ & $52.05$ & $-0.54 \pm 0.36$ \\
\midrule
E2E    & pooled                   & ---     & ---     & $-0.14 \pm 0.37$ \\
SAMSum & pooled                   & ---     & ---     & $+0.00 \pm 0.37$ \\
\bottomrule
\end{tabular}
\end{table}

Neither method recovers held-out gradient signs on adapter tensors: absolute
accuracies lie between $48.3\%$ and $52.6\%$, that is, within a few points of
chance, and every difference is smaller than its seed standard deviation. The
pooled effect is $-0.14 \pm 0.37$ points on E2E and $+0.00 \pm 0.37$ on SAMSum.
We read this as a null result rather than as weak evidence in either direction,
and note that where \ZEMA{} governs only a thin adapter atop a frozen backbone,
it achieves nothing measurable. Sign recovery on LoRA adapters should not be
offered as evidence for the method.

\subsection{Optimizers from other families: Muon, GaLore, and ZEMA inside GaLore}
\label{sec:other-families}

The arms in this appendix are not coordinate-wise sign methods, so they do not
belong in Table~\ref{tab:pretraining}; but they are the two best-known ways to
train a transformer on less than two optimizer states per parameter, and leaving
them out would make the memory claim in this paper look less contested than it
is. They run in the identical setting: same architecture, corpus, schedule,
batch, decay and three seeds.

All three use the same hybrid split \ZEMA{} uses. Only the $84.9$M body linear
weights ($52.4\%$ of the model) go to the arm; the embedding, the untied head
and the norms stay on AdamW, because Muon and GaLore both require matrices and
neither the reference implementations nor ours applies them to an embedding
table. State totals therefore include an AdamW pair on the remaining $77.2$M
parameters.

\begin{table}[h]
\centering
\caption{Optimizers from other families, same setting as
Table~\ref{tab:pretraining}: final validation perplexity after $3.2$B tokens,
mean of three seeds, $95\%$ Student-$t$ interval on the per-seed paired
difference. ``Body state'' counts optimizer coordinates over the $84.9$M body
weights; ``total'' adds the AdamW pair carried on the remaining $77.2$M
parameters. AdamW and \ZEMA{}-CLT are repeated from Table~\ref{tab:pretraining}
for reference.}
\label{tab:other-families}
\small
\begin{tabular}{lccccc}
\toprule
Optimizer & lr & body state & total state & val.\ ppl & $\Delta$ vs.\ AdamW \\
\midrule
AdamW & $8{\cdot}10^{-3}$ & $169.9$M & $324.4$M & $22.79 \pm 0.46$ & --- \\
\ZEMA{}-CLT & $8{\cdot}10^{-3}$ & $84.9$M & $239.4$M & $22.18 \pm 0.12$ & $-0.61 \pm 0.55$ \\
\midrule
Muon & $3{\cdot}10^{-2}$ & $84.9$M & $239.4$M & $\mathbf{20.14 \pm 0.11}$ & $-2.65 \pm 0.40$ \\
GaLore & $1{\cdot}10^{-2}$ & $73.1$M & $227.6$M & $23.58 \pm 0.22$ & $+0.79 \pm 0.28$ \\
\ZEMA{}-CLT/GaLore & $1{\cdot}10^{-2}$ & $\mathbf{44.8}$M & $\mathbf{199.3}$M & $24.10 \pm 0.07$ & $+1.31 \pm 0.52$ \\
\ZEMA{}-Adam/GaLore & $1{\cdot}10^{-2}$ & $44.8$M & $199.3$M & $24.14 \pm 0.09$ & $+1.36 \pm 0.54$ \\
\bottomrule
\end{tabular}
\end{table}

\begin{figure}[h]
\centering
\includegraphics[width=0.82\textwidth]{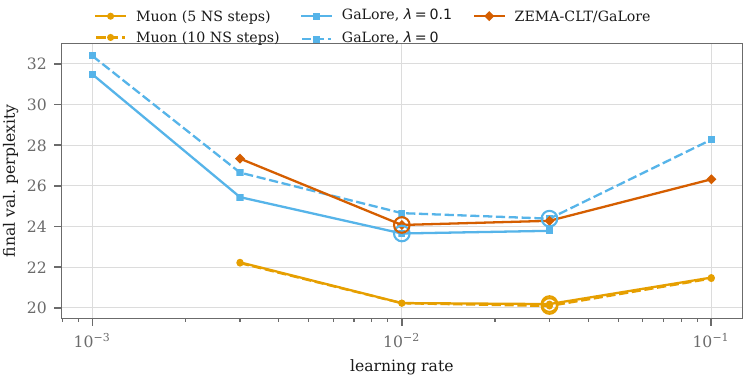}
\caption{Pretraining sweeps for the arms of Table~\ref{tab:other-families};
the ring marks the selected rate. Every argmin is interior to its own grid. As
in Figure~\ref{fig:pretrain-sweep-baselines}, a variant of an optimizer keeps
that optimizer's colour and is separated by the dash pattern: Muon's two
Newton--Schulz depths lie within $0.10$ perplexity of each other at every rate,
which is why the five-step reference configuration is the one reported, and
GaLore's two weight decays select different rates. \ZEMA{}-CLT/GaLore carries
\ZEMA{}-CLT's colour from Table~\ref{tab:pretraining} because it is that arm, run
inside GaLore's subspace; it tracks GaLore-Adam's curve shape but sits above it
at every rate, including at the $10^{-2}$ both arms select.}
\label{fig:pretrain-sweep-families}
\end{figure}

\paragraph{Muon.} One momentum buffer per matrix, whose direction is replaced by
its nearest semi-orthogonal matrix through a five-step Newton--Schulz iteration
in bfloat16, at the reference coefficients. It is the same one-state budget as
\ZEMA{} and it wins this comparison outright, by $2.65$ perplexity against AdamW,
four times \ZEMA{}'s margin. We report this plainly because it bounds what the rest
of the paper claims: a single state per parameter is clearly sufficient to beat
Adam at this scale, and \ZEMA{} is not the only way to get there. What \ZEMA{} and
Muon do is different in kind---Muon changes the geometry the step is measured
in, and is defined only for matrices; \ZEMA{} changes the per-coordinate scale, and
applies to any parameter---and nothing in our results says the two cannot be
combined. Its rate was swept on $\{3\cdot10^{-3}, 10^{-2}, 3\cdot10^{-2},
10^{-1}\}$, selecting $3\cdot10^{-2}$ (perplexities $22.2$, $20.2$, $\ast20.2$,
$21.5$; Figure~\ref{fig:pretrain-sweep-families}), inside the published
$0.02$--$0.05$ band. A ten-step Newton--Schulz
variant was also run and finished within $0.10$ of the five-step arm at every
rate, so the five-step reference configuration is what we report.

\paragraph{GaLore.} Adam run inside a rank-$r$ subspace of each gradient, taken
from that gradient's own SVD and refreshed every $200$ steps. At $r=256$ its
state is $r(\min + 2\max)$ per matrix, which for a $768\times768$ matrix is
exactly $d^2$: the published rank for $d_{\text{model}}=768$ and the rank at
which GaLore matches \ZEMA{}'s state coincide, so this is GaLore at its own
setting, not squeezed to fit our comparison. It trails AdamW by $0.79$ and \ZEMA{}
by $1.40$, at $0.86\times$ \ZEMA{}'s body state. We swept both weight decays:
$\lambda=0.1$ (matched to every other arm) selects $10^{-2}$ and is the row
above; the reference's own pretraining default $\lambda=0$ selects a different
rate, $3\cdot10^{-2}$, and is worse by $0.85$ at $24.45 \pm 0.15$. Running only
the published default would have made GaLore the sole unregularized arm here and
cost it most of a perplexity point.

\paragraph{\ZEMA{} inside GaLore.} The two memory mechanisms compose in principle:
GaLore keeps three states of a smaller shape, \ZEMA{} keeps one state of full
shape, and running \ZEMA{} as the inner optimizer keeps one state of the smaller
shape. This is the cheapest arm we ran, at $0.53\times$ \ZEMA{}'s body state and
$0.26\times$ AdamW's. It is cheap to compute as well as to store: projecting a
per-example gradient requires projecting only one of its two operands, since
$G_i = \delta_i^\top x_i$ gives $G_i P^\top = \delta_i^\top (x_i P^\top)$ for a
tall matrix and $P^\top G_i = (\delta_i P)^\top x_i$ for a wide one, so \ZEMA{}'s
per-example reduction is unchanged and simply runs at the smaller shape.

\emph{It does not compose in practice.} Against GaLore-Adam at the same rate,
decay and subspace---the comparison in which only the inner optimizer
changes---the swap costs $+0.52 \pm 0.29$ (CLT) and $+0.57 \pm 0.28$ ($t_2$),
paired over three seeds. The rate is not the explanation: we swept it over
$\{3\cdot10^{-3}, 10^{-2}, 3\cdot10^{-2}, 10^{-1}\}$
(Figure~\ref{fig:pretrain-sweep-families}) and both arms optimize at
the same $10^{-2}$, with the swap costing $0.42$--$0.49$ at every rate near the
optimum and $1.90$ at a rate $3\times$ too small.

Two observations point the same way. \ZEMA{}'s advantage inverts under projection:
at full shape \ZEMA{}-CLT beats AdamW by $0.61$, inside a rank-$256$ subspace it
loses to Adam by $0.52$. And the choice of link, which separates the two \ZEMA{}
variants by $0.18$ at full shape, stops mattering inside the subspace---the
paired difference between them there is $+0.045 \pm 0.061$, an interval
containing zero. Both are consistent with the projection having already
discarded the per-coordinate structure that \ZEMA{}'s standardization exploits: the
top-$r$ directions of the gradient are not where the two links disagree.

\section{Training horizon in TinyStories pretraining}
\label{app:horizon}

The comparison in Section~\ref{sec:pretraining} is made at a single token
budget. This appendix asks how much the outcome depends on that budget, in a
second and much smaller pretraining setting: a $9.31$M-parameter model on
TinyStories, trained to $3.2$B tokens at two batch sizes with three seeds per
arm. The comparison there has three phases. AdamW leads early, \ZEMA{} leads
through the middle of training, and the two finish level, so a run stopped
early can report either optimizer ahead. The same configuration stopped at a
fixed update count also produces an apparent growth of AdamW's margin with
batch size, which the full horizon shows to be an effect of where training
stops rather than of batch size.

\subsection{Setup}
\label{app:horizon-setup}

The model is a $9.31$M-parameter decoder-only transformer (6 layers,
$d_{\mathrm{model}} = 256$, 8 heads, FFN width 768, context 128) trained on
TinyStories with a native $8{,}192$-token BPE vocabulary, with exact
within-batch dispersion, body-only \ZEMA{} coverage, and a micro-batch of 64
sequences. Each run trains for $3.2$B tokens under a cosine schedule with
$10\%$ linear warmup over the full horizon, and is evaluated every $0.1$B
tokens on $4$M held-out tokens, giving 32 checkpoints per run. Validation
batches are drawn from a generator reseeded identically at every checkpoint, so
all checkpoints of all runs are scored on the same held-out sequences and the
training stream is unaffected by the evaluation cadence.

We run $B = 256$ and $B = 1024$ with three seeds per optimizer, twelve runs
total. Learning rates are the bracketed optima of a six-point grid, searched
separately for each optimizer at $2{,}000$ updates at that batch: $0.006$ at
$B = 256$ and $0.012$ at $B = 1024$. At both batches the two optimizers
selected the same value, so a single learning rate is simultaneously each one's
own optimum and neither arm is advantaged by the choice. Runs used one NVIDIA
H100 each on Trillium.

\subsection{Results}
\label{app:horizon-results}

\begin{table}[t]
\centering
\caption{Validation perplexity against token budget, three seeds per cell.
$\Delta$ is \ZEMA{}-CLT minus AdamW at matched seed, with a two-sided $95\%$
$t$-interval over the three paired differences; negative favours \ZEMA{}. The
final column counts seeds favouring \ZEMA{}. Bold marks intervals excluding zero.}
\label{tab:horizon}
\begin{tabular}{lrrrrc}
\toprule
Tokens & AdamW & \ZEMA{}-CLT & $\Delta$ & $95\%$ CI & seeds \\
\midrule
\multicolumn{6}{l}{\emph{$B = 256$, lr $0.006$}} \\
$0.10$B & $7.2566$ & $7.3466$ & $\mathbf{+0.0900}$ & $[+0.036, +0.144]$ & 0/3 \\
$0.20$B & $6.6933$ & $6.7274$ & $+0.0341$ & $[-0.002, +0.070]$ & 0/3 \\
$0.40$B & $6.4488$ & $6.4458$ & $-0.0030$ & $[-0.044, +0.038]$ & 2/3 \\
$0.80$B & $6.2208$ & $6.1947$ & $-0.0261$ & $[-0.083, +0.031]$ & 3/3 \\
$1.20$B & $6.0241$ & $5.9959$ & $\mathbf{-0.0282}$ & $[-0.055, -0.002]$ & 3/3 \\
$1.60$B & $5.7759$ & $5.7559$ & $\mathbf{-0.0200}$ & $[-0.035, -0.005]$ & 3/3 \\
$2.00$B & $5.5002$ & $5.4793$ & $-0.0209$ & $[-0.050, +0.008]$ & 3/3 \\
$2.40$B & $5.1610$ & $5.1492$ & $-0.0117$ & $[-0.038, +0.015]$ & 2/3 \\
$2.80$B & $4.7832$ & $4.7736$ & $-0.0095$ & $[-0.021, +0.002]$ & 3/3 \\
$3.20$B & $4.5885$ & $4.5850$ & $-0.0035$ & $[-0.015, +0.008]$ & 2/3 \\
\midrule
\multicolumn{6}{l}{\emph{$B = 1024$, lr $0.012$}} \\
$0.10$B & $7.3837$ & $7.6183$ & $\mathbf{+0.2346}$ & $[+0.152, +0.317]$ & 0/3 \\
$0.20$B & $6.5267$ & $6.6138$ & $\mathbf{+0.0871}$ & $[+0.029, +0.146]$ & 0/3 \\
$0.40$B & $6.0843$ & $6.0821$ & $-0.0022$ & $[-0.032, +0.027]$ & 1/3 \\
$0.80$B & $5.7894$ & $5.7766$ & $-0.0127$ & $[-0.056, +0.030]$ & 3/3 \\
$1.20$B & $5.6291$ & $5.6011$ & $-0.0280$ & $[-0.064, +0.008]$ & 3/3 \\
$1.60$B & $5.4257$ & $5.3933$ & $\mathbf{-0.0324}$ & $[-0.046, -0.019]$ & 3/3 \\
$2.00$B & $5.1954$ & $5.1694$ & $\mathbf{-0.0260}$ & $[-0.041, -0.011]$ & 3/3 \\
$2.40$B & $4.9196$ & $4.9021$ & $\mathbf{-0.0175}$ & $[-0.031, -0.005]$ & 3/3 \\
$2.80$B & $4.6519$ & $4.6357$ & $-0.0162$ & $[-0.046, +0.014]$ & 3/3 \\
$3.20$B & $4.5319$ & $4.5208$ & $-0.0112$ & $[-0.030, +0.008]$ & 3/3 \\
\bottomrule
\end{tabular}
\end{table}

Table~\ref{tab:horizon} shows the same three-phase structure at both batch
sizes. AdamW leads early by a margin that is significant at $0.10$B and larger
at the larger batch ($+0.235$ versus $+0.090$). The curves cross at
approximately $0.4$B tokens. \ZEMA{} then leads for the remainder of training,
by $0.01$ to $0.03$ perplexity, with all three seeds agreeing at every
checkpoint from $0.8$B onward and several intervals excluding zero. As the
cosine anneals the gap closes, and at $3.2$B the two are level: $-0.004$
$[-0.015, +0.008]$ at $B = 256$ and $-0.011$ $[-0.030, +0.008]$ at $B = 1024$,
neither separable from zero.

Two points follow for short-horizon comparisons. First, measured at $3.2$B
tokens the difference is $-0.1\%$ at $B = 256$ and $-0.2\%$ at $B = 1024$,
whereas the same configuration stopped at $2{,}000$ optimizer
updates---$0.065$B tokens at $B = 256$ and $0.26$B at $B = 1024$---puts AdamW
ahead by $2.09\%$ and $3.56\%$. Second, a fixed update budget also generates an
apparent batch-size trend: AdamW's early advantage is itself larger at the
larger batch, and every cell is sampled inside that early window, so a horizon
effect is read off as a batch effect. At the full horizon the trend does not
survive, because there is no advantage to either method.

The crossover is determined by tokens rather than by updates. It occurs near
$0.4$B tokens at both batch sizes, which is $12{,}207$ optimizer steps at
$B = 256$ but only $3{,}052$ at $B = 1024$. Any explanation resting on
optimizer-state warm-up would predict a fixed step count; the data instead
tracks data seen, consistent with \ZEMA{}'s evidence weighting becoming useful
once gradient noise grows relative to the gradient mean.

Finally, note that $B = 1024$ reaches a lower perplexity than $B = 256$ at
matched tokens for both optimizers ($4.53$ and $4.52$ versus $4.59$ for both),
so the larger batch is not harmful in this setting once training is long
enough.

\subsection{Smoothing rule ablation}
\label{app:horizon-smoothing}

The belief EMA in Algorithm~\ref{alg:zema} applies an update of much smaller
magnitude than the per-step belief it smooths: at $B = 256$ the recorded mean
absolute belief is $0.50$--$0.83$ while the mean absolute applied update is
$0.07$--$0.085$, a factor of $6.5$--$9.8$, and the ratio is largest early in
training. This invites the hypothesis that the shrinkage is a defect and that
correcting it would remove \ZEMA{}'s early disadvantage. It does not.
Table~\ref{tab:smoothing} compares the four alternative smoothing rules
implemented alongside the paper's, each with its own learning rate selected
from a common six-point grid spanning $2.5\times10^{-4}$ to $8\times10^{-3}$ at
$B = 256$ and $2{,}000$ updates.

\begin{table}[t]
\centering
\caption{Smoothing rule ablation, $B = 256$, $2{,}000$ updates, seed 7.
Each variant's learning rate is selected independently; all optima are interior
to the grid searched.}
\label{tab:smoothing}
\begin{tabular}{llrl}
\toprule
Variant & Rule & Val.\ PPL & lr \\
\midrule
AdamW & --- & $6.2607$ & $0.006$ \\
\ZEMA{}-CLT & \textbf{EMA of the belief (Algorithm~\ref{alg:zema})} & $\mathbf{6.3917}$ & $0.006$ \\
\ZEMA{}-CLT-z & EMA of $z$, then link & $7.1721$ & $0.002$ \\
\ZEMA{}-pool & pooled $z$, bias corrected by $\sqrt{1-\beta^{2t}}$ & $7.3559$ & $0.001$ \\
\ZEMA{}-logodds & accumulate log-odds, squash with $\tanh$ & $7.5795$ & $0.001$ \\
\ZEMA{}-kalman & posterior mean, bias corrected by $1-\beta^{t}$ & $8.0148$ & $0.0005$ \\
\bottomrule
\end{tabular}
\end{table}

Every alternative is worse, and the ordering is monotone in how aggressively
the variant removes the shrinkage: the selected learning rate falls from
$0.006$ to $0.0005$ across the table as the effective update grows, and
perplexity degrades in step, with the most thorough corrector worst by $1.6$
perplexity. The damping is therefore doing useful work. Averaging bounded
beliefs cancels coordinates whose sign disagrees between steps, and the
variants that undo it trade that evidence weighting for larger, noisier
updates. We keep the post-link EMA of Algorithm~\ref{alg:zema} on this
evidence.

\subsection{Caveats}
\label{app:horizon-caveats}

Intermediate checkpoints are mid-schedule: the cosine is tied to the full
$3.2$B horizon, so the value at $0.4$B in these runs is not what a run
terminating at $0.4$B with a fully annealed learning rate would produce. The
comparison between arms at each checkpoint is unaffected, since both share the
schedule, but these intermediate numbers are not comparable to annealed
endpoints. The learning rates were selected at $2{,}000$ updates and reused at
$97{,}656$; both arms carry the same value at each batch, so the comparison
remains fair, but absolute perplexities may be off-optimum. The $3.2$B horizon
is roughly $7.7$ epochs over the $416$M-token training split; validation
perplexity is still decreasing at the final checkpoint, so neither optimizer
has been trained to saturation. Finally, perplexity is specific to the
$8{,}192$-token vocabulary used here and is not comparable to published
TinyStories perplexities obtained with different tokenizers.

\clearpage
\section{Efficient implementation of \ZEMA{}}
\label{app:efficient-implementation}

Section~\ref{sec:pretraining} reports \ZEMA{} at a per-step cost of $1.046\times$
AdamW. A direct implementation of Algorithm~\ref{alg:zema} costs $1.59\times$.
This appendix specifies the three changes that close that gap and the
end-to-end runs that validate them. Two of the three are exact rearrangements
of the same computation; the third replaces one estimator of
$\boldsymbol{\sigma}_t$ with another estimator of the same quantity, at a cost
of three degrees of freedom out of $255$. Section~\ref{sec:dispersion-details}
then treats subsampling, which is an independent and larger reduction available
on top of these.

\subsection{Where the overhead comes from}
\label{sec:overhead-anatomy}

Fix one bound linear weight of shape $d_{\text{out}} \times d_{\text{in}}$, and
let $L$ be the context length, $B$ the global batch of sequences, and $W$ the
number of data-parallel ranks. For example $n$ write
$\boldsymbol{x}^{(n)} \in \mathbb{R}^{L \times d_{\text{in}}}$ for the layer
input, $\boldsymbol{\delta}^{(n)} \in \mathbb{R}^{L \times d_{\text{out}}}$ for
the gradient of the loss with respect to the layer output, and
\begin{equation}
\label{eq:per-example-gradient}
\boldsymbol{g}^{(n)} = \boldsymbol{\delta}^{(n)\top}\boldsymbol{x}^{(n)}
  \in \mathbb{R}^{d_{\text{out}} \times d_{\text{in}}}
\end{equation}
for that example's own weight gradient. Algorithm~\ref{alg:zema} needs the mean
$\boldsymbol{m}_t$ and, through line~\ref{line:zema-standard-error}, the
dispersion
\begin{equation}
\label{eq:dispersion-sum}
\boldsymbol{M}_{2} = \sum_{n=1}^{B}\bigl(\boldsymbol{g}^{(n)}-\boldsymbol{m}_{t}\bigr)^{2},
\qquad
\boldsymbol{\sigma}_{t}^{2} = \frac{\boldsymbol{M}_{2}}{B(B-1)} .
\end{equation}

\paragraph{The asymmetry.} $\boldsymbol{m}_t$ is the ordinary weight gradient
that any backward pass already produces, and costs nothing extra.
$\boldsymbol{M}_2$ is a function of the spread \emph{across} examples, which the
usual batched backward destroys the instant it sums over the batch. Every cost
below follows from that one fact, and it appears twice. In arithmetic, AdamW's
weight-gradient GEMM contracts over all $B{\cdot}L$ tokens at once, whereas any
per-example statistic must stop at every example boundary. In communication,
AdamW's second moment $\boldsymbol{v}_t = \beta_2\boldsymbol{v}_{t-1} +
(1-\beta_2)\boldsymbol{g}_t^{2}$ is a pure function of the \emph{already
reduced} gradient, which every rank holds after the gradient all-reduce, so it
costs no extra bytes; $\boldsymbol{M}_2$ is a function of the
\emph{pre-reduction} spread, which that same reduction has destroyed.

\paragraph{The direct implementation.} Written literally,
Equation~\ref{eq:dispersion-sum} materializes the
$[B, d_{\text{out}}, d_{\text{in}}]$ per-example gradient tensor, writes it once
and reads it at least twice; and because $\boldsymbol{M}_2$ is defined against
the \emph{global} mean, it forces three serialized collectives per step---reduce
the mean, agree a common scale, then reduce the rescaled dispersion---of which
the second cannot begin until the first has finished. Table~\ref{tab:fast-ladder}
gives the resulting ladder.

\begin{table}[h]
\centering
\caption{Cumulative effect of the changes in this appendix, on the setting of
Section~\ref{sec:pretraining}. Each row is a complete $6{,}103$-step run at seed
$1$. Times are training only; the separate validation pass, $156$--$158$\,s in
every row, is excluded. The perplexities carry no interval of their own and should be
read against the three-seed spread of $\pm 0.049$ reported for this arm in
Table~\ref{tab:pretraining}; the numerical changes are gated separately in
Table~\ref{tab:fast-gates}. $^{\dagger}$These runs were launched to gate the
numerical changes, so the final row is the configuration current at that time:
it additionally carried the variant discarded in \S\ref{sec:overlap} and
predates the host-side cleanup described in \S\ref{sec:fast-validation}. It is
therefore not the configuration of Table~\ref{tab:fast-throughput}.}
\label{tab:fast-ladder}
\begin{tabular}{lrrr}
\toprule
Configuration & Train time & Tokens/s & vs.\ AdamW \\
\midrule
AdamW & $3.572$ h & $248{,}815$ & $1.000\times$ \\
\midrule
Direct implementation of Algorithm~\ref{alg:zema} & $5.643$ h & $157{,}496$ & $1.580\times$ \\
\quad $+$ \texttt{bfloat16} statistic accumulation & $4.828$ h & $184{,}103$ & $1.352\times$ \\
\quad $+$ rank-local pooling (\S\ref{sec:rank-local}) & $4.322$ h & $205{,}644$ & $1.210\times$ \\
\quad $+$ overlapped reduction (\S\ref{sec:overlap}) \\
\quad\quad and fused per-example kernel (\S\ref{sec:example-kernel})$^{\dagger}$ & $3.862$ h & $230{,}129$ & $1.081\times$ \\
\bottomrule
\end{tabular}
\end{table}

\subsection{Rank-local pooling of the dispersion}
\label{sec:rank-local}

\paragraph{The change.} Pool the dispersion within each rank only. Partition the
batch into the $W$ rank-sized blocks it is already sliced into, write
$\bar{\boldsymbol{g}}_{w}$ for the mean over the $B/W$ examples resident on rank
$w$, and decompose the exact sum of squares:
\begin{equation}
\label{eq:anova}
\underbrace{\sum_{n}\bigl(\boldsymbol{g}^{(n)}-\boldsymbol{m}_{t}\bigr)^{2}}_{\text{total}}
\;=\;
\underbrace{\sum_{w}\sum_{n \in w}\bigl(\boldsymbol{g}^{(n)}-\bar{\boldsymbol{g}}_{w}\bigr)^{2}}_{\text{within ranks}}
\;+\;
\underbrace{\frac{B}{W}\sum_{w}\bigl(\bar{\boldsymbol{g}}_{w}-\boldsymbol{m}_{t}\bigr)^{2}}_{\text{between ranks}} .
\end{equation}
If examples are i.i.d.\ across ranks, the three terms have expectations
$(B-1)\varsigma^{2}$, $(B-W)\varsigma^{2}$ and $(W-1)\varsigma^{2}$ for a
per-example variance $\varsigma^{2}$. Discarding the between-rank term and
dividing by $B-W$ rather than $B-1$ therefore leaves an unbiased estimator of
the \emph{same} quantity:
\begin{equation}
\label{eq:rank-local}
\sigma_{t,i}^{2}
  \;=\; \frac{1}{B(B-W)}\sum_{w}\sum_{n \in w}\bigl(g^{(n)}_{i}-\bar{g}_{w,i}\bigr)^{2} .
\end{equation}
As with subsampling (Equation~\ref{eq:dispersion-subsampling}), the estimand is
unchanged---it is still the standard error of the $B$-example mean---and
$\boldsymbol{m}_t$ itself is untouched, still the exact global mean formed by an
all-reduce over every example. Only the sample used for the dispersion changes,
and with it the relative error of $\boldsymbol{\sigma}_t$, which for a
$\chi^{2}$ estimator on $\nu$ degrees of freedom is $1/\sqrt{2\nu}$:

\begin{center}
\begin{tabular}{lrr}
\toprule
Estimator of $\boldsymbol{\sigma}_t$ & $\nu$ & rel.\ error \\
\midrule
Exact pooling (Algorithm~\ref{alg:zema}) & $255$ & $4.43\%$ \\
Rank-local pooling, $W=4$ & $252$ & $4.45\%$ \\
Subsampling at $k=32$ (\S\ref{sec:dispersion-details}) & $31$ & $12.70\%$ \\
\bottomrule
\end{tabular}
\end{center}

\paragraph{Why it is cheaper.} With no between-rank term, nothing in the
dispersion depends on the pooled mean, so the three-deep chain collapses to two
independent sums:
\begin{align*}
\text{exact:}\quad
  & \textsc{sum}(\text{means})
    \;\to\; \textsc{max}(\text{scale})
    \;\to\; \textsc{sum}(\text{rescaled }\boldsymbol{M}_{2}) , \\
\text{rank-local:}\quad
  & \textsc{sum}(\boldsymbol{m}) \;\Vert\; \textsc{sum}(\boldsymbol{M}_{2}) ,
\end{align*}
where $\Vert$ denotes two collectives issued back to back with no dependency
between them.
Because each rank's window holds only $B/W$ examples, the partials can further
be carried as raw sums $\boldsymbol{S}_1 = \sum_{n\in w}\boldsymbol{g}^{(n)}$ and
$\boldsymbol{S}_2 = \sum_{n \in w}(\boldsymbol{g}^{(n)})^{2}$, converted locally
by $\boldsymbol{M}_2 = \boldsymbol{S}_2 - \boldsymbol{S}_1^{2}W/B$ with no
collective at all, which removes the \textsc{max} round that the factored form
needs for underflow protection. The cancellation this invites is bounded by the
identity $S_2 = M_2\,(1 + z^{2}/(B/W-1))$: at $z_{\max}=10$ and $B/W = 64$ the
worst attainable ratio is $2.6$, under one bit of \texttt{float32}, and at the
$|z| \approx 0.51$ these runs actually exhibit it is $1.004$.

\paragraph{Assumptions, and when they fail.} Two conditions are load-bearing.
First, unbiasedness in Equation~\ref{eq:rank-local} requires the between-rank
term to be pure noise. Here the global permutation is drawn once per step and
\emph{then} sliced across ranks, so rank membership carries no information and
the condition holds by construction; an implementation that instead gave each
rank its own shard, or sorted examples by length, would discard real
between-rank spread, report $\boldsymbol{\sigma}_t$ low, $|\boldsymbol{z}_t|$
high, and be overconfident. Second, the raw-sum form assumes no underflow. It is
safe at pretraining scale---the fraction of coordinates reporting zero variance
is exactly $0$ on every run reported here---but not universally: LoRA
initializes one factor at zero, producing gradients of order $10^{-22}$ whose
squares underflow \texttt{float32} to exactly zero and read as unanimous
agreement among examples. The fine-tuning experiments of
Section~\ref{sec:finetuning} therefore retain the factored form and its
\textsc{max} round.

\subsection{A fused per-example kernel}
\label{sec:example-kernel}

\paragraph{Why the ordinary GEMM structure does not apply.} A matrix-product
kernel tiles the output $[d_{\text{out}}, d_{\text{in}}]$ and streams the whole
$B{\cdot}L$ contraction into a register accumulator that is written once. The
per-example square in Equation~\ref{eq:dispersion-sum} requires the sum over the
$L$ token positions to \emph{complete} at every example boundary, so the
accumulator must be finalized $B$ times rather than once---which is why the
direct path falls back to a batched product and materializes
$[B, d_{\text{out}}, d_{\text{in}}]$.

\paragraph{The change.} Reorder the contraction so that the example index is the
outer loop and finalize in registers, never writing the per-example tensor
(Algorithm~\ref{alg:zema-kernel}). Each program owns a disjoint tile of the
output, so the final accumulation is a plain read--modify--write rather than an
atomic. Operands are loaded in their own precision with a \texttt{float32}
accumulator---the same arithmetic the ordinary weight gradient performs---so
tensor cores are retained; upcasting both operands before the product instead
forces the \texttt{float32} CUDA-core path and is $4.7$--$6.5\times$ slower than
the plain weight gradient.

\begin{algorithm2e}
\caption{Fused per-example first and second sums, for one bound weight and one
rank. Replaces line~\ref{line:zema-per-example-gradients} of
Algorithm~\ref{alg:zema}; $\boldsymbol{S}_1$ is the rank's weight gradient and
$\boldsymbol{S}_2$ feeds Equation~\ref{eq:rank-local}. The
$[B,d_{\text{out}},d_{\text{in}}]$ tensor is never formed.}
\label{alg:zema-kernel}
\SetAlgoLined
\LinesNumbered
\DontPrintSemicolon
\KwIn{activations $\boldsymbol{x}$, output sensitivities $\boldsymbol{\delta}$;
      tile sizes $(b_{\text{out}}, b_{\text{in}}, b_{L})$}
\KwOut{$\boldsymbol{S}_1, \boldsymbol{S}_2$}
\ForEach{tile $\mathcal{T}$ of size $b_{\text{out}} \times b_{\text{in}}$ of the output, in parallel}{
  $\boldsymbol{s}_1 \leftarrow \mathbf{0}$; $\boldsymbol{s}_2 \leftarrow \mathbf{0}$
    \tcp*{\texttt{float32}, in registers}
  \For{$n \leftarrow 1$ \KwTo $B$}{
    $\boldsymbol{p} \leftarrow \mathbf{0}$
      \tcp*{this example's own gradient tile}
    \For{each block of $b_L$ token positions}{
      $\boldsymbol{p} \leftarrow \boldsymbol{p}
        + \boldsymbol{\delta}^{(n)\top}_{\text{block}}\boldsymbol{x}^{(n)}_{\text{block}}$
        \tcp*{tensor-core product}
    }
    $\boldsymbol{s}_1 \leftarrow \boldsymbol{s}_1 + \boldsymbol{p}$;\quad
    $\boldsymbol{s}_2 \leftarrow \boldsymbol{s}_2 + \boldsymbol{p}^{2}$
      \tcp*{finalize at the example boundary}
  }
  write $\boldsymbol{s}_1, \boldsymbol{s}_2$ into $\mathcal{T}$ of $\boldsymbol{S}_1, \boldsymbol{S}_2$\;
}
\end{algorithm2e}

\paragraph{It replaces work rather than only adding it.} $\boldsymbol{S}_1$
\emph{is} the weight gradient, so the kernel subsumes the backward pass's own
weight-gradient product for that layer, which must then be suppressed. On the
setting of Section~\ref{sec:pretraining} the kernel costs $181$\,ms of device
time per step while removing a $148$\,ms product, a marginal cost of about
$33$\,ms; an accounting that prices it as pure addition overstates it roughly
fivefold. The extra arithmetic proper is $B\,d_{\text{out}}d_{\text{in}}$
squares against $B\,L\,d_{\text{out}}d_{\text{in}}$ multiply--accumulates,
smaller by a factor of $L$. The binding constraint is registers: three
\texttt{float32} tiles are live at once, which caps tile size well below what a
plain GEMM would choose.

\paragraph{When it pays.} The saving is the traffic of the
$[B, d_{\text{out}}, d_{\text{in}}]$ tensor, weighed against operands that are
read either way, so the governing ratio is
\begin{equation}
\label{eq:kernel-payoff}
r \;=\; \frac{B\,d_{\text{out}}d_{\text{in}}}{B\,L\,(d_{\text{out}}+d_{\text{in}})}
  \;=\; \frac{d_{\text{out}}d_{\text{in}}}{L\,(d_{\text{out}}+d_{\text{in}})} .
\end{equation}
Here $r = 0.19$--$0.27$, but $r > 1$ whenever the context is short and the layer
wide---$1.94$ for a $[8192, 256]$ head at $L=128$. Short context and wide layers
favour the kernel; long context and narrow layers do not.

\subsection{Overlapping the reduction with the backward pass}
\label{sec:overlap}

With gradient accumulation, a weight's $\boldsymbol{S}_1$ and $\boldsymbol{S}_2$
are complete only once the \emph{last} micro-batch's backward has passed it.
Reducing everything after the accumulation window closes leaves the collective
fully exposed at the end of the step. But weights become final in reverse layer
order, so the reduction for the last layer can be launched while the earlier
layers are still computing. Grouping weights into byte-sized buckets and
launching each bucket asynchronously as it fills---the scheme
\texttt{DistributedDataParallel} uses for gradients---recovers $59$\,ms per step
here. It changes only when bytes move, not what is computed.

Two conditions are required. Bucket boundaries must be identical on every rank,
or ranks issue mismatched collectives and deadlock; this holds because weights
complete in the same deterministic order everywhere and buckets fill by the same
byte rule. And the scheme requires rank-local pooling: under the exact
three-round chain of \S\ref{sec:rank-local}, round two depends on round one and
round three on round two, so only the first could be launched early. Removing
the between-rank term is what makes both payloads independent, and therefore
launchable as soon as they exist.

\paragraph{A variant we discarded.} One can go further and use the
\emph{previous} step's dispersion with the current step's mean, leaving the
current dispersion in flight across the step boundary. We do not recommend it.
Once the kernel and the bucketing above had shrunk the surrounding work,
bucketed overlap alone hid the dispersion collective essentially completely, and
the stale variant measured $-1$\,ms---within noise---while costing a persistent
$[d_{\text{out}}, d_{\text{in}}]$ \texttt{float32} buffer, $+0.32$\,GiB, which is
precisely the footprint of AdamW's $\boldsymbol{v}$. \ZEMA{}'s claim is one state
against Adam's two; that variant buys a second full-size state for no speed.

\subsection{What remains, and why it cannot reach zero}
\label{sec:comm-floor}

After all three changes, \ZEMA{} moves two weight-shaped payloads per step on the
bound parameters where AdamW moves one:

\begin{center}
\begin{tabular}{lrr}
\toprule
Per step, bound parameters & Bytes & vs.\ AdamW \\
\midrule
AdamW (gradient only) & $648{,}735{,}744$ & $1.00\times$ \\
\ZEMA{}, rank-local (mean $+$ dispersion $+$ auxiliary) & $988{,}474{,}368$ & $1.52\times$ \\
\bottomrule
\end{tabular}
\end{center}

One of the two is not a premium: the mean \emph{is} the gradient, the same bytes
data-parallel training moves anyway. The second has no AdamW counterpart and
cannot acquire one, for the reason given in \S\ref{sec:overhead-anatomy}:
$\boldsymbol{v}_t$ is recoverable from the reduced gradient and
$\boldsymbol{M}_2$ is not. $1.52\times$ of AdamW's collective traffic is
therefore the floor for this statistic as specified, not a target to optimize
past. Reducing it further means computing a different statistic---which is what
subsampling (\S\ref{sec:dispersion-details}) does, and it is priced in degrees
of freedom accordingly.

\subsection{Validation}
\label{sec:fast-validation}

Of the four changes in Table~\ref{tab:fast-ladder}, two alter the numbers the
optimizer sees---\texttt{bfloat16} accumulation of the statistic, through
rounding, and rank-local pooling, through three degrees of freedom---and two do
not. The fused kernel computes Equation~\ref{eq:per-example-gradient} exactly in
exact arithmetic and is pinned against a batched-product reference in unit
tests; bucketed overlap changes only scheduling.

Both numerical changes were carried to a \emph{complete} run---$6{,}103$ steps
and $3.2$B tokens, ending in a full validation pass over the $9.2$M held-out
tokens---rather than to a short smoke, and compared against the exact baseline
with a tolerance of $\pm 2\varsigma$, where $\varsigma = 0.002189$ nats is that
baseline's own three-seed standard deviation.

\begin{table}[h]
\centering
\caption{End-to-end validation of the numerical changes, on the setting of
Section~\ref{sec:pretraining}, seed $1$. Tolerance $\pm 2\varsigma = \pm
0.004378$ nats.}
\label{tab:fast-gates}
\begin{tabular}{lrrrc}
\toprule
Configuration & Val.\ loss & $\Delta$ & Perplexity & Verdict \\
\midrule
Exact baseline & $3.096886$ & --- & $22.129$ & --- \\
$+$ \texttt{bfloat16} statistic & $3.099870$ & $+0.002984$ & $22.195$ & pass \\
$+$ rank-local pooling & $3.100192$ & $+0.003306$ & $22.202$ & pass \\
$+$ fused kernel & $3.096976$ & $+0.000090$ & $22.131$ & pass \\
\bottomrule
\end{tabular}
\end{table}

\paragraph{Short runs are not a substitute.} A rounding change and a
three-degree-of-freedom change both produce drift that is invisible early and
only interpretable against a same-schedule multi-seed band, which exists only at
the end of training. A $20$-step smoke of this configuration reported a
negative-variance fraction of $0.142$ where the completed runs give
$3.9\cdot10^{-9}$, because the smoke's compressed warmup pooled $8$ examples
where the real schedule pools $256$.

\paragraph{Measured throughput.} Table~\ref{tab:fast-throughput} gives per-step
training times for the two endpoints. AdamW's own step time varies by about
$7\%$ across nodes of this cluster, so each block is one job on one node and
ratios are taken within a block, never across. Peak memory is \emph{below}
AdamW's in every \ZEMA{} row, because \ZEMA{} carries one persistent state
tensor on the bound parameters where AdamW carries two.

\paragraph{Reconciling the two ratios.} Table~\ref{tab:fast-ladder} ends at
$1.081\times$ and Table~\ref{tab:fast-throughput} at $1.046\times$. Both are
training-time measurements and the difference is configuration, not metric. The
complete runs of Table~\ref{tab:fast-ladder} exist to gate the two numerical
changes, and were launched before the implementation settled: that final row
still carries the stale-dispersion variant of \S\ref{sec:overlap} and predates
a set of host-side cleanups---chiefly the removal of $672$ per-step
device-to-host synchronizations---together worth $90$\,ms per step, taking
$2{,}278$\,ms down to the $2{,}188$\,ms of
Table~\ref{tab:fast-throughput}. The configuration we recommend has been
measured per step but not carried to a complete run. We therefore report
$1.046\times$ as its per-step cost and $1.081\times$ as the only ratio backed
by a full $3.2$B-token run, rather than quoting one number for both.

\begin{table}[h]
\centering
\caption{Per-step training time and peak memory, four L40S, global batch $256$
at context $2048$. Each block is a single job on a single node; the two blocks
are separate jobs, which is why AdamW appears twice. The middle row of the
second block isolates the $59$\,ms recovered by overlapping the reduction
(\S\ref{sec:overlap}).}
\label{tab:fast-throughput}
\begin{tabular}{lrrr}
\toprule
Configuration & ms/step & vs.\ AdamW & Peak GiB \\
\midrule
AdamW & $2109$ & $1.000\times$ & $22.20$ \\
Direct implementation of Algorithm~\ref{alg:zema} & $3359$ & $1.593\times$ & $21.79$ \\
\midrule
AdamW & $2091$ & $1.000\times$ & $22.20$ \\
Rank-local pooling $+$ fused kernel, no overlap & $2247$ & $1.075\times$ & $21.15$ \\
All three changes & $2188$ & $1.046\times$ & $21.15$ \\
\bottomrule
\end{tabular}
\end{table}

\paragraph{Diagnostics for a reimplementation.} Three quantities are worth
monitoring. The fraction of coordinates reporting zero variance must be $0$; a
nonzero value means the raw-sum cancellation bound of \S\ref{sec:rank-local} has
failed for that regime and the factored form is required. The mean $|z|$, about
$0.51$ here, sets the realized cancellation ratio and should be checked before
trusting the raw-sum path. And the assignment of examples to ranks must be
verified to be exchangeable, or Equation~\ref{eq:rank-local} is biased rather
than merely wider. Kernel tile sizes are register-bound and shape-dependent, so
they should be retuned per architecture---after computing
Equation~\ref{eq:kernel-payoff}, which decides whether the kernel is worth using
at all.

\subsection{Subsampling the per-batch variance}

\label{sec:dispersion-details}

\begin{figure}[h]
\centering
\includegraphics[width=0.66\textwidth]{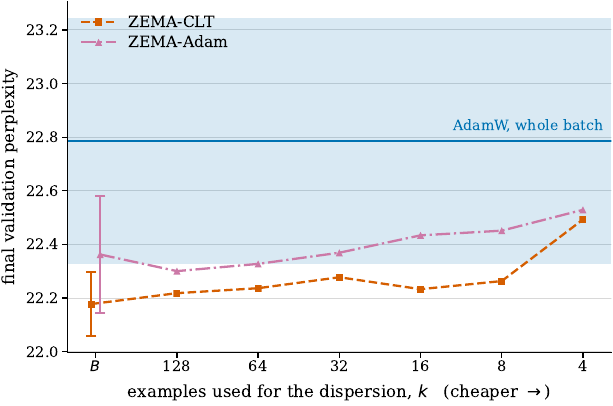}
\caption{Estimating the variance from $k$ of the $B=256$ examples, on the
setting of Section~\ref{sec:pretraining}. Left to right is dearer to cheaper.
The leftmost point uses every example and is the three-seed mean of
Table~\ref{tab:pretraining} with its $95\%$ interval; every other point is a
single seed and so carries no bar. The AdamW line and band are its own
whole-batch three-seed mean and $95\%$ interval. Signum, at $25.25$, is off the
top of the scale.}
\label{fig:dispersion}
\end{figure}

The standard error $\boldsymbol{\sigma}_t$ is the only statistic in \ZEMA{} that
needs per-example gradients, and it does not need all of them: the mean can be
taken over the whole batch while the variance is estimated from a fraction of
it. Figure~\ref{fig:dispersion} sweeps that fraction on the setting of
Section~\ref{sec:pretraining}. Down to $k=8$ of $B=256$ examples both arms stay
within seed noise of their whole-batch result, and at every $k$ we tried they
remain below AdamW. Details of the estimator and the
Algorithm~\ref{alg:zema-subsampled} can be found below.

\paragraph{The estimator.} The mean $\boldsymbol{m}_t$ and the standard error
$\boldsymbol{\sigma}_t$ need not be estimated from the same examples. We keep the
mean over all $B$ examples and estimate the dispersion from a subset $S$ of size
$k$:
\begin{equation}
\label{eq:dispersion-subsampling}
\sigma_{t,i}^{2}
  = \frac{1}{B(k-1)}\sum_{j \in S}\bigl(g_{j,i} - \bar{g}_{S,i}\bigr)^{2},
\qquad
\bar{g}_{S,i} = \frac{1}{k}\sum_{j \in S} g_{j,i}.
\end{equation}
The divisor is $B(k-1)$, not $k(k-1)$. The sum over $S$ divided by $k-1$
estimates the \emph{per-example} variance from $k$ draws; dividing by $B$ instead
rescales that to the standard error of the $B$-example mean, which is the
quantity $\boldsymbol{z}_t$ is defined against. The estimand is therefore
unchanged---only the sample used to estimate it shrinks, and with it the relative
error of $\boldsymbol{\sigma}_t$, which grows as roughly $1/\sqrt{2(k-1)}$.

\textbf{This is not a smaller batch.} Every example still enters
$\boldsymbol{m}_t$, so the direction the optimizer follows is formed from the
full batch, and the $n$ behind the standard error remains $B$. A batch of $k$
would change both.

\paragraph{Why it is cheaper.} Algorithm~\ref{alg:zema-subsampled} gives the
replacement. The saving is in traffic, not arithmetic. The full path materializes
a $[B, \text{out}, \text{in}]$ per-example gradient tensor and makes several
elementwise passes over it; here the mean comes from a single matrix product over
all $B{\cdot}T$ tokens---the same contraction, and the same floating-point
operations, as the ordinary weight gradient---and only $k$ examples ever enter
the batched product that forms per-example gradients. Arithmetic \emph{rises} by
a factor $(B+k)/B$; the intermediate tensor shrinks by $B/k$.

\begin{algorithm2e}
\caption{Subsampled variance. Replaces
lines~\ref{line:zema-per-example-gradients}--\ref{line:zema-standard-error} of
Algorithm~\ref{alg:zema}; everything from $\boldsymbol{z}_t$ onwards is
unchanged.}
\label{alg:zema-subsampled}
\SetAlgoLined
\LinesNumbered
\DontPrintSemicolon
\KwIn{batch size $B$; variance sample size $k \leq B$; activations
      $\boldsymbol{x}$ and output sensitivities $\boldsymbol{\delta}$ for the
      batch}
sample a batch of size $B$\;
$\boldsymbol{m}_{t} \leftarrow \frac{1}{B}\,\boldsymbol{\delta}^{\top}\boldsymbol{x}$
  \tcp*{one product over all $B{\cdot}T$ tokens; no per-example tensor}
$S \leftarrow$ the first $k$ examples of the batch
  \tcp*{the batch is shuffled, so $S$ is a random draw from the data}
form per-example gradients $\boldsymbol{g}^{(j)}_{t}$ for $j \in S$ \emph{only}
  \tcp*{a $[k,\text{out},\text{in}]$ tensor, not $[B,\text{out},\text{in}]$}
$\bar{\boldsymbol{g}}_{S} \leftarrow \frac{1}{k}\sum_{j \in S} \boldsymbol{g}^{(j)}_{t}$
  \tcp*{centre on the subsample's own mean}
$\boldsymbol{\sigma}_{t} \leftarrow
  \bigl[\frac{1}{B(k-1)}\sum_{j \in S}(\boldsymbol{g}^{(j)}_{t}-\bar{\boldsymbol{g}}_{S})^{2}\bigr]^{1/2}$
  \tcp*{still the standard error of the $B$-example mean}
\end{algorithm2e}

\paragraph{Results.} Figure~\ref{fig:dispersion} sweeps $k$ from $256$ down to
$4$, one seed per $k$. From $k=128$ down to $k=8$---a $32\times$ reduction in the
per-example tensor---\ZEMA{}-CLT moves between $22.22$ and $22.28$ against its
whole-batch $22.18$, and \ZEMA{}-Adam between $22.30$ and $22.45$ against $22.36$;
in both cases the excursion is at or below the seed-to-seed spread at that point,
which is $0.05$ and $0.09$ in standard deviation over the three whole-batch
seeds. Only at $k=4$ does a penalty appear that is clearly bigger than noise, and
it remains small: \ZEMA{}-CLT rises to $22.49$ and \ZEMA{}-Adam to $22.53$. At every
$k$, including $k=4$, both arms sit below AdamW's whole-batch mean of $22.79$,
though at $k=4$ they are inside AdamW's $95\%$ interval and we do not claim a
separation there.

\end{document}